\documentclass[authordate,onecolumn]{miri-tech-article}

\usepackage{miritools}
\usepackage[utf8]{inputenc}
\usepackage[english]{babel}
\usepackage{csquotes}
\usepackage[export]{adjustbox} 

\definecolor{r}{RGB}{197, 46, 14}
\definecolor{o}{RGB}{197, 119, 14}
\definecolor{b}{RGB}{45, 14, 197}
\definecolor{g}{RGB}{92, 180, 32}

\providecommand{\tightlist}{%
  \setlength{\itemsep}{2pt}\setlength{\parskip}{2pt}
  }

\usepackage{lipsum}
\usepackage[notextcomp]{kpfonts} 

\title{Embedded Agency}

\author{Abram Demski \and Scott Garrabrant \\ Machine Intelligence Research Institute \\ \{abram,scott\}@intelligence.org}

\begin{document}

\nocopyright

\maketitle

\begin{abstract}
    Traditional models of rational action treat the agent as though it is cleanly separated from its environment, and can act on that environment from the outside. Such agents have a known functional relationship with their environment, can model their environment in every detail, and do not need to reason about themselves or their internal parts.

    We provide an informal survey of obstacles to formalizing good reasoning for agents \emph{embedded} in their environment. Such agents must optimize an environment that is not of type ``function''; they must rely on models that fit within the modeled environment; and they must reason about themselves as just another physical system, made of parts that can be modified and that can work at cross purposes.
\end{abstract}

\tableofcontents
\newpage

\section{Introduction}\label{sec:ea}

Suppose that you want to design an AI system to achieve some real-world goal---a goal that requires the system to learn for itself and figure out many things that you don't already know.

There's a complicated engineering problem here. But there's also a problem of figuring out what it even means to build a learning agent like that. What is it to optimize realistic goals in physical environments? In broad terms, how does it work? 

In this article, we'll point to four ways we \emph{don't} currently know how it works, and four areas of active research aimed at figuring it out.

\vspace{4mm}

\subsection{Embedded agents}

\label{ea1} This is Alexei, and Alexei is playing a video game.

\vspace{6mm}

\begin{center}
  \includegraphics[width=0.9\textwidth, cfbox=gray]{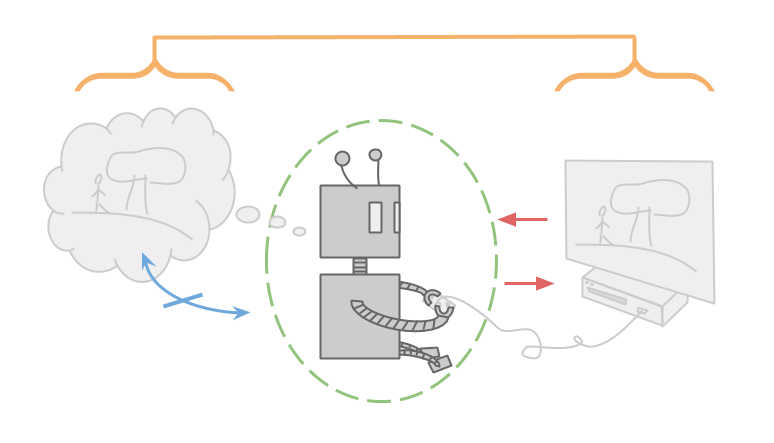}
\end{center}

\vspace{6mm}

\noindent Like most games, this game has \textcolor{r}{clear input and output channels}. Alexei only observes the game through the computer screen, and only manipulates the game through the controller. The game can be thought of as a function which takes in a sequence of button presses and outputs a sequence of pixels on the screen. 

Alexei is also very smart, and capable of \textcolor{o}{holding the entire video game inside his mind}. If Alexei has any uncertainty, it is only over empirical facts like what game he is playing, and not over logical facts like which inputs (for a given deterministic game) will yield which outputs. This means that Alexei must also store inside his mind every possible game he could be playing. 

Alexei \textcolor{b}{does not, however, have to think about himself}. He is only optimizing the game he is playing, and not optimizing the brain he is using to think about the game. He may still choose actions based off of value of information, but this is only to help him rule out possible games he is playing, and not to change the way in which he thinks. 

In fact, Alexei can treat himself as \textcolor{g}{an unchanging indivisible atom}. Since he doesn't exist in the environment he's thinking about, Alexei doesn't worry about whether he'll change over time, or about any subroutines he might have to run. 

Notice that all the properties we talked about are partially made possible by the fact that Alexei is cleanly separated from the environment that he is optimizing. 

\newpage

\label{ea2} This is Emmy. Emmy is playing real life. 

\vspace{3mm}

\begin{center}
    \includegraphics[width=1.0\textwidth, cfbox=gray]{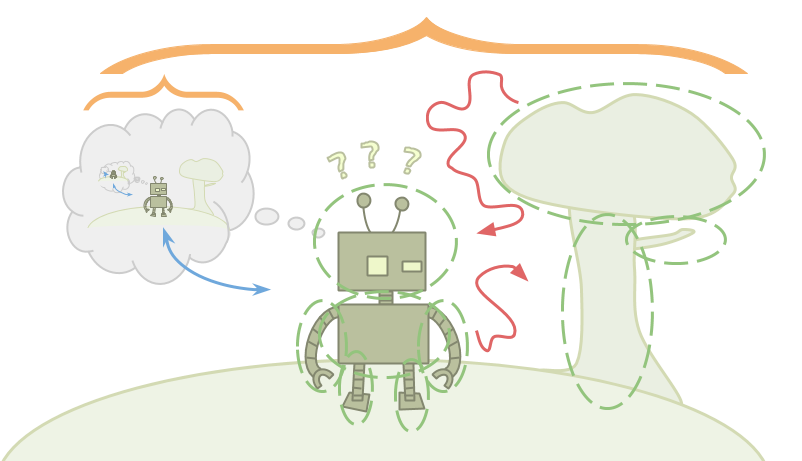}
\end{center}

\vspace{6mm}

\noindent Real life is not like a video game. The differences largely come from the fact that Emmy is within the environment that she is trying to optimize. 

Alexei sees the universe as a function, and he optimizes by choosing inputs to that function that lead to greater reward than any of the other possible inputs he might choose. Emmy, on the other hand, \textcolor{r}{doesn't have a function}. She just has an environment, and this environment contains her. 

Emmy wants to choose the best possible action, but which action Emmy chooses to take is just another fact about the environment. Emmy can reason about the part of the environment that is her decision, but since there's only one action that Emmy ends up actually taking, it's not clear what it even means for Emmy to ``choose'' an action that is better than the rest. 

Alexei can poke the universe and see what happens. Emmy is the universe poking itself. In Emmy's case, how do we formalize the idea of ``choosing'' at all? 

To make matters worse, since Emmy is contained within the environment, Emmy must also be \textcolor{o}{smaller than the environment}. This means that Emmy is incapable of storing accurate detailed models of the environment within her mind. 

This causes a problem: Bayesian reasoning works by starting with a large collection of possible environments, and as you observe facts that are inconsistent with some of those environments, you rule them out. What does reasoning look like when you're not even capable of storing a single valid hypothesis for the way the world works? Emmy is going to have to use a different type of reasoning, and make updates that don't fit into the standard Bayesian framework. 

Since Emmy is within the environment that she is manipulating, she is also going to be capable of \textcolor{b}{self-improvement}. But how can Emmy be sure that as she learns more and finds more and more ways to improve herself, she only changes herself in ways that are actually helpful? How can she be sure that she won't modify her original goals in undesirable ways? 

Finally, since Emmy is contained within the environment, she can't treat herself like an atom. She is \textcolor{g}{made out of the same pieces} that the rest of the environment is made out of, which is what causes her to be able to think about herself. 

In addition to hazards in her external environment, Emmy is going to have to worry about threats coming from within. While optimizing, Emmy might spin up other optimizers as subroutines, either intentionally or unintentionally. These subsystems can cause problems if they get too powerful and are unaligned with Emmy's goals. Emmy must figure out how to reason without spinning up intelligent subsystems, or otherwise figure out how to keep them weak, contained, or aligned fully with her goals. 

\vspace{4mm}

\label{ea3} Emmy is confusing, so let's go back to Alexei. Marcus Hutter's \citeyearpar{Hutter:2005:uai,Hutter:2012:decadeuai} \textcolor{r}{AIXI} framework gives a good theoretical model for how agents like Alexei work:

$$
  a_k \;:=\; \arg\max_{a_k}\sum_{ o_k r_k} 
   ... \max_{a_m}\sum_{ o_m r_m}
  [r_k+...+r_m]
\hspace{-1em}\hspace{-1em}\hspace{-1em}\!\!\!\sum_{{ q}\,:\,U({ q},{ a_1..a_m})={ o_1 r_1.. o_m r_m}}\hspace{-1em}\hspace{-1em}\hspace{-1em}\!\!\! 2^{-\ell({ q})}
$$ 
The model has an agent and an environment that interact using actions, observations, and rewards. The agent sends out an action $a$, and then the environment sends out both an observation $o$ and a reward $r$. This process repeats at each time $k\ldots{}m$. 

Each action is a function of all the previous action-observation-reward triples. And each observation and reward is similarly a function of these triples and the immediately preceding action. 

You can imagine an agent in this framework that has full knowledge of the environment that it's interacting with. However, AIXI is used to model optimization under uncertainty about the environment. AIXI has a distribution over all possible computable environments $q$, and chooses actions that lead to a high expected reward under this distribution. Since it also cares about future reward, this may lead to exploring for value of information. 

Under some assumptions, we can show that AIXI does reasonably well in all computable environments, in spite of its uncertainty. However, while the environments that AIXI is interacting with are computable, AIXI itself is uncomputable. The agent is made out of a different sort of stuff, a more powerful sort of stuff, than the environment. 

We will call agents like AIXI and Alexei ``dualistic''. They exist outside of their environment, with \textcolor{r}{only set interactions between agent-stuff and environment-stuff}. They \textcolor{o}{require the agent to be larger than the environment}, and \textcolor{b}{don't tend to model self-referential reasoning}, because \textcolor{g}{the agent is made of different stuff than what the agent reasons about}.

AIXI is not alone. These dualistic assumptions show up all over our current best theories of rational agency.

\label{ea4} We set up AIXI as a bit of a foil, but AIXI can also be used as inspiration. When the authors look at AIXI, we feel like we really understand how Alexei works. This is the kind of understanding that we want to also have for Emmy.

Unfortunately, Emmy is confusing. When we talk about wanting to have a theory of ``embedded agency,'' we mean that we want to be able to understand theoretically how agents like Emmy work. That is, agents that are embedded within their environment and thus: 

\begin{itemize}
 \tightlist
  \item \textcolor{r}{do not have well-defined i/o channels;}
  \item \textcolor{o}{are smaller than their environment;}
  \item \textcolor{b}{are able to reason about themselves and self-improve;}
  \item \textcolor{g}{and are made of parts similar to the environment.}
\end{itemize}

\noindent You shouldn't think of these four complications as a partition. They are very entangled with each other. 

For example, the reason the agent is able to self-improve is because it is made of parts. And any time the environment is sufficiently larger than the agent, it might contain other copies of the agent, and thus destroy any well-defined i/o channels.

\begin{center}
  \makebox[\textwidth][c]{\includegraphics[width=1.3\textwidth]{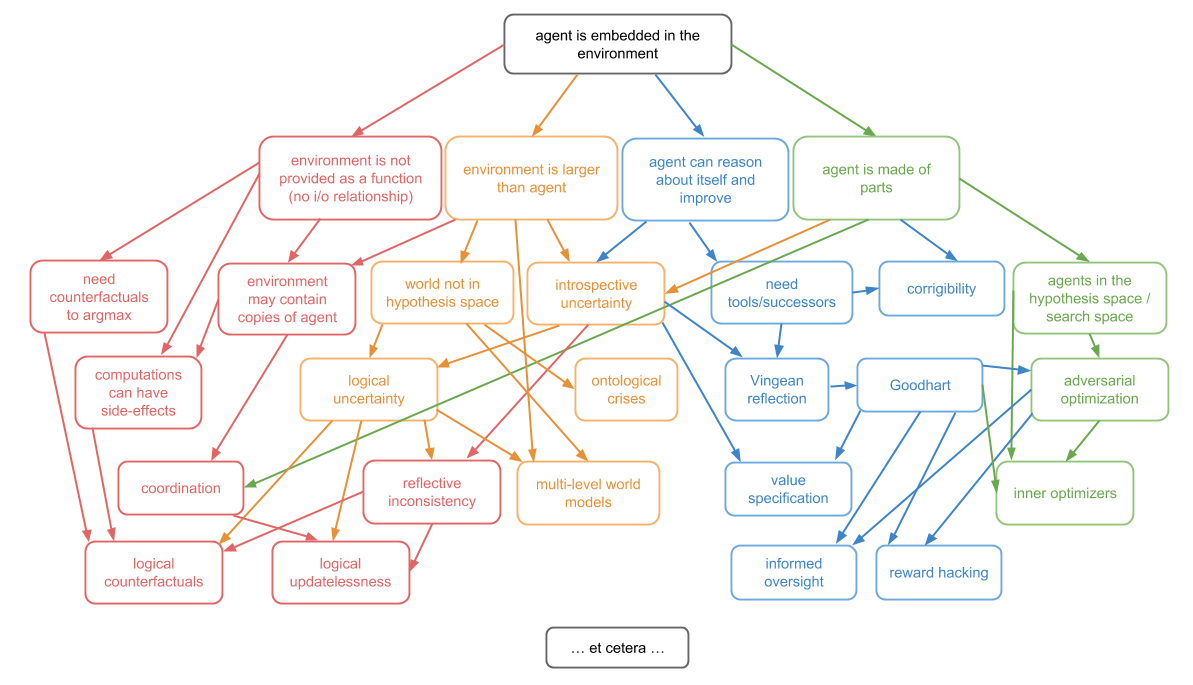}} 
\end{center}

\vspace{3mm}

\noindent However, we will use these four complications to inspire a split of the topic of embedded agency into four subproblems. These are: \textcolor{r}{decision theory}, \textcolor{o}{embedded world-models}, \textcolor{b}{robust delegation}, and \textcolor{g}{subsystem alignment}.

\subsection{Short problem descriptions}
\label{ea5}\label{ea:dt}\textcolor{r}{\textbf{Decision theory}} is about embedded optimization. 

The simplest model of dualistic optimization is $\argmax$. $\argmax$ takes in a function from actions to rewards, and returns the action which leads to the highest reward under this function.

Most optimization can be thought of as some variant on this. You have some space; you have a function from this space to some score, like a reward or utility; and you want to choose an input that scores highly under this function.

But we just said that a large part of what it means to be an embedded agent is that we don't have a functional environment. How does optimization work in that case? Optimization is clearly an important part of agency, but we can't currently say what it is even in theory without making major type errors.

Some major open problems in decision theory include: 

\begin{itemize}
 \tightlist
  \item \textbf{logical counterfactuals}: how do you reason about what would happen if you take action B, given that you can prove that you will instead take action A?
  \item environments that include multiple \textbf{copies of the agent}, or trustworthy predictions of the agent.
  \item \textbf{logical updatelessness}, which is about how to combine the very nice but very \emph{Bayesian} world of Wei Dai's (\citeyear{Dai:2009:udt}) updateless decision theory, with the much less Bayesian world of logical uncertainty.
\end{itemize}

\noindent \label{ea:ewm}\textcolor{o}{\textbf{Embedded world-models}} is about how to make good models of the world that are able to fit within an agent that is much smaller than the world. 

This has proven to be very difficult---first, because it means that the true universe is not in the agent's hypothesis space, which ruins a number of theoretical guarantees; and second, because it means the agent must make non-Bayesian updates as it learns, which also ruins many theoretical guarantees. 

It is also about how to make world-models from the point of view of an observer on the inside, and resulting problems such as anthropics. Some major open problems in embedded world-models include: 

\begin{itemize}
 \tightlist
  \item \textbf{logical uncertainty}, which is about how to combine the world of logic with the world of probability.
  \item \textbf{multi-level modeling}, which is about how to have multiple models of the same world at different levels of description, and transition nicely between them.
  \item \textbf{ontological crises}, which is what to do when you realize that your model, or even your goal, was specified using a different ontology than the real world.
\end{itemize}

\noindent \textcolor{b}{\label{ea:rd}\textbf{Robust delegation}} is about a special type of principal-agent problem. You have an initial agent that wants to make a more intelligent successor agent to help it optimize its goals. The initial agent has all of the power, because it gets to decide exactly what successor agent to make. But in another sense, the successor agent has all of the power, because it is much, much more intelligent. 

From the point of view of the initial agent, the question is about creating a successor that will robustly not use its intelligence against you. From the point of view of the successor agent, the question is, ``How do you robustly learn or respect the goals of something that is stupid, manipulable, and not even using the right ontology?'' And there are additional problems coming from the \emph{L\"{o}bian obstacle} making it impossible to consistently trust reasoning systems that are more powerful than you.

You can think about these problems in the context of an agent that is learning over time, or in the context of an agent making a significant self-improvement, or in the context of an agent that is just trying to make a powerful tool.

The major open problems in robust delegation include: 

\begin{itemize}
 \tightlist
  \item \textbf{Vingean reflection}, which is about how to reason about and trust agents that are much smarter than you, in spite of the L\"{o}bian obstacle to trust.
  \item \textbf{value learning}, the successor agent's task of learning the goals of the initial agent in spite of the initial agent's inconsistencies and lower intelligence.
  \item \textbf{corrigibility}, which is about how an initial agent can get a successor agent to allow (or even help with) modifications, in spite of an instrumental incentive not to \citep{Soares:2015:corrigibility}.
\end{itemize}

\noindent \label{ea:sa}\textcolor{g}{\textbf{Subsystem alignment}} is about how to be \emph{one unified agent} that doesn't have subsystems that are fighting against either you or each other. 

When an agent has a goal, like ``saving the world,'' it might end up spending a large amount of its time thinking about a subgoal, like ``making money''. If the agent spins up a sub-agent that is only trying to make money, there are now two agents that have different goals, and this leads to a conflict. The sub-agent might suggest plans that look like they \emph{only} make money, but actually destroy the world in order to make even more money. 

The problem is that we do not just need to worry about sub-agents that we intentionally spin up. We also have to worry about spinning up sub-agents by accident. Any time we perform a search or an optimization over a sufficiently rich space that is able to contain agents, we have to worry about the space itself doing optimization. This optimization may not be exactly in line with the optimization the outer system was trying to do, but it \emph{will} have an instrumental incentive to \emph{look} as though it is aligned. 

A lot of optimization in practice uses this kind of passing the buck. You don't just find a solution; you find a thing that is able to itself search for a solution. 

In theory, we don't understand how to do \emph{optimization} at all---other than methods that look like finding a bunch of stuff that we don't understand, and seeing if it accomplishes our goal. But this is exactly the kind of thing that's most prone to spinning up adversarial subsystems.

The big open problem in subsystem alignment is about how to have a base-level optimizer that doesn't spin up adversarial optimizers. You can break this problem up further by considering cases where the resultant optimizers are either \textbf{intentional} or \textbf{unintentional}, and considering restricted subclasses of optimization, like \textbf{induction}.

We describe each of these problems in more detail below, while attempting to keep in view that decision theory, embedded world-models, robust delegation, and subsystem alignment are not four separate problems. They are different subproblems of the same unified concept of \emph{embedded agency}.

\section{Decision theory}\label{sec:dt}

Decision theory and artificial intelligence typically try to compute something resembling 

$$\underset{a \ \in \ Actions}{\argmax} \ f(a).$$

\vspace{3mm}

\noindent I.e., maximize some function of the action. This tends to assume that we can disentangle things enough to see outcomes as a function of actions. For example, AIXI represents the agent and the environment as separate units which interact over time through clearly defined i/o channels, so that it can then choose actions maximizing reward. 

When the agent model is a part of the environment model, it can be significantly less clear how to consider taking alternative actions.

\label{dt1} For example, because the agent is \textcolor{o}{smaller than the environment}, there can be other copies of the agent, or things very similar to the agent. This leads to contentious decision-theory problems such as the twin prisoner's dilemma and Newcomb's problem \citep{Nozick:1969,Gibbard:1978}. If Emmy Model 1 and Emmy Model 2 have had the same experiences and are running the same source code, should Emmy Model 1 act like her decisions are steering both robots at once? Depending on how you draw the boundary around ``yourself'', you might think you control the action of both copies, or only your own.\footnote{For a general introduction to this topic, see \citet{Yudkowsky:2017:fdt}.}

This is an instance of the problem of counterfactual reasoning: how do we evaluate hypotheticals like “What if the sun suddenly went out”?

\label{dt2} Problems of adapting \textcolor{r}{\textbf{decision theory}} to embedded agents include: 

\begin{itemize}
 \tightlist
  \item counterfactuals
  \item Newcomblike reasoning, in which the agent interacts with copies of itself
  \item reasoning about other agents more broadly
  \item extortion problems
  \item coordination problems
  \item logical counterfactuals
  \item logical updatelessness
\end{itemize}

\subsection{Action counterfactuals}
\noindent \label{dt3} The most central example of why agents need to think about counterfactuals comes from counterfactuals about their own actions.

The difficulty with action counterfactuals can be illustrated by the five-and-ten problem \citep{Fallenstein:2013:510,Garrabrant:2017:obstacles}. Suppose we have the option of taking a five dollar bill or a ten dollar bill, and all we care about in the situation is how much money we get. Obviously, we should take the \$10.

However, it is not so easy as it seems to reliably take the \$10. If you reason about yourself as just another part of the environment, then you can \textcolor{b}{know your own behavior}. If you can know your own behavior, then it becomes difficult to reason about what would happen if you behaved differently.

This throws a monkey wrench into many common reasoning methods. How do we formalize the idea ``Taking the \$10 would lead to good consequences, while taking the \$5 would lead to bad consequences,'' when sufficiently rich self-knowledge would reveal one of those scenarios as inconsistent? Or if we can't formalize any idea like that, how do real-world agents nonetheless figure out to take the \$10?

If we try to calculate the expected utility of our actions by Bayesian conditioning, as is common, knowing our own behavior leads to a divide-by-zero error when we try to calculate the expected utility of actions we know we don't take: $\lnot A$ implies $P(A)=0$, which implies $P(B \& A)=0$, which implies

$$P(B|A) = \frac{P(B \& A)}{P(A)} = \frac{0}{0}.$$

\noindent Because the agent doesn't know how to separate itself from the environment, it gets gnashing internal gears when it tries to imagine taking different actions.

But the biggest complication comes from \textcolor{b}{L\"{o}b's theorem} \citep{Lob:1955},\footnote{See \citet{LaVictoire:2015:lobintro} for a discussion of L\"{o}b's theorem as it relates to embedded agency.} which can make otherwise reasonable-looking agents take the \$5 because ``If I take the \$10, I get \$0''! Moreover, this error turns out to be stable—the problem can't be solved by the agent learning or thinking about the problem more.

This might be hard to believe; so let’s look at a detailed example. The phenomenon can be illustrated by the behavior of simple logic-based agents reasoning about the five-and-ten problem.

Consider this example:

~
\vspace{2mm}

\SetKwProg{blank}{}{} 

\begin{algorithm}[H]
\blank{$A:=$}
{Spend some time searching for proofs of sentences of the form ``$[A()=5 \to U()=x] \ \ \& \ \ [A()=10 \to U()=y]$''  for $x,y \in \{0,5,10\}$.\;
 \eIf{a proof is found with $x>y$}{
  \Return 5\;
  }{
  \Return 10\;
 }
}
\caption{Agent}
\end{algorithm}

~
\vspace{2mm}

\begin{algorithm}[H]
\blank{$U:=$}
{
\If{$A()=10$}
 {
  \Return 10\;
  }
\If{$A()=5$}
 {
  \Return 5\;
  }
 }
\caption{Universe}
\end{algorithm}

~
\vspace{4mm}

\noindent We have the source code for an agent and the universe. They can refer to each other through the use of quining. The universe is simple; the universe just outputs whatever the agent outputs. 

The agent spends a long time searching for proofs about what happens if it takes various actions. If for some $x$ and $y$ equal to $0$, $5$, or $10$, it finds a proof that taking the $5$ leads to $x$ utility, that taking the $10$ leads to $y$ utility, and that $x>y$, it will naturally take the $5$. We expect that it won't find such a proof, and will instead pick the default action of taking the $10$. 

It seems easy when you just imagine an agent trying to reason about the universe. Yet it turns out that if the amount of time spent searching for proofs is enough, the agent will always choose $5$! 

The proof that this is so is by L\"{o}b's theorem. L\"{o}b's theorem says that, for any proposition $P$, if you can prove that a \emph{proof} of $P$ would imply the \emph{truth} of $P$, then you can prove $P$. In symbols, with ``$\square X$'' meaning ``$X$ is provable'': 

$$\square (\square P \to P) \to \square P.$$ 

\noindent In the version of the five-and-ten problem we gave, ``$P$'' is the proposition ``if the agent outputs $5$ the universe outputs $5$, and if the agent outputs $10$ the universe outputs $0$''. 

Supposing it is provable, the agent will eventually find the proof, and return $5$ in fact. This makes the sentence \emph{true}, since the agent outputs $5$ and the universe outputs $5$, and since it's false that the agent outputs $10$. This is because false propositions like ``the agent outputs $10$'' imply everything, \emph{including} the universe outputting $5$. 

The agent can (given enough time) prove all of this, in which case the agent in fact proves the proposition ``if the agent outputs $5$ the universe outputs $5$, and if the agent outputs $10$ the universe outputs $0$''. And as a result, the agent takes the \$5. 

We call this a ``spurious proof'': the agent takes the \$5 because it can prove that \emph{if} it takes the \$10 it has low value, \emph{because} it takes the \$5. It sounds circular, but sadly, is logically correct. More generally, when working in less proof-based settings, we refer to this as a problem of spurious counterfactuals. 

The general pattern is: counterfactuals may spuriously mark an action as not being very good. This makes the AI not take the action. Depending on how the counterfactuals work, this may remove any feedback which would ``correct'' the problematic counterfactual; or, as we saw with proof-based reasoning, it may actively help the spurious counterfactual be ``true''. 

Note that because the proof-based examples are of significant interest to us, ``counterfactuals'' actually have to be \textbf{counter\emph{logicals}}; we sometimes need to reason about logically impossible ``possibilities''. This rules out most existing accounts of counterfactual reasoning. 

\label{dt5} You may have noticed that we slightly cheated. The only thing that broke the symmetry and caused the agent to take the \$5 was the fact that ``$5$'' was the action that was taken when a proof was found, and ``$10$'' was the default. We could instead consider an agent that looks for any proof at all about what actions lead to what utilities, and then takes the action that is better. This way, which action is taken is dependent on what order we search for proofs. 

Let's assume we search for short proofs first. In this case, we will take the \$10, since it is very easy to show that $A()=5$ leads to $U()=5$ and $A()=10$ leads to $U()=10$. 

The problem is that spurious proofs can be short too, and don't get much longer when the universe gets harder to predict. If we replace the universe with one that is provably functionally the same, but is harder to predict, the shortest proof will short-circuit the complicated universe and be spurious. 

People often try to solve the problem of counterfactuals by suggesting that there will always be some uncertainty. An AI may know its source code perfectly, but it can't perfectly know the hardware it is running on. 

Does adding a little uncertainty solve the problem? Often not: 

\begin{itemize}
 \tightlist
  \item The proof of the spurious counterfactual often still goes through; if you think you are in a five-and-ten problem with a 95\% certainty, you can have the usual problem within that 95\%.
  \item Adding uncertainty to make counterfactuals well-defined doesn't get you any guarantee that the counterfactuals will be \emph{reasonable}. Hardware failures aren't often what you want to expect when considering alternate actions.
\end{itemize}

\noindent Consider this scenario: You are confident that you almost always take the left path. However, it is possible (though unlikely) for a cosmic ray to damage your circuits, in which case you could go right---but you would then be insane, which would have many other bad consequences \citep{Bensinger:2017:badoutcomes}.

If \emph{this reasoning in itself} is why you always go left, you've gone wrong. 

Simply ensuring that the agent has some uncertainty about its actions does not ensure that the agent will have remotely reasonable counterfactual expectations. However, one thing we can try instead is to ensure the agent actually takes each action with some probability. This strategy is called \textbf{epsilon-exploration}.

Epsilon-exploration ensures that if an agent plays similar games on enough occasions, it can eventually learn realistic counterfactuals (modulo a concern of \textcolor{o}{realizability} which we will get to later).

Epsilon-exploration only works if it ensures that the agent itself can't predict whether it is about to epsilon-explore. In fact, a good way to implement epsilon-exploration is via the rule ``if the agent is too sure about its action, it takes a different one.''

From a logical perspective, the unpredictability of epsilon-exploration is what prevents the problems we've been discussing. From a learning-theoretic perspective, if the agent could know it wasn't about to explore, then it could treat that as a different case---failing to generalize lessons from its exploration. This gets us back to a situation where we have no guarantee that the agent will learn better counterfactuals. Exploration may be the only source of data for some actions, so we need to force the agent to take that data into account, or it may not learn.

However, even epsilon-exploration does not seem to get things exactly right. Observing the result of epsilon-exploration shows you what happens if you take an action \emph{unpredictably}; the consequences of taking that action as part of business-as-usual may be different.

Suppose you are an epsilon-explorer who lives in a world of epsilon-explorers. You are applying for a job as a security guard, and you need to convince the interviewer that you are not the kind of person who would steal the objects you are guarding. The interviewer wants to hire someone who has too much integrity to lie and steal, even if the person thought that they could get away with it.

Suppose that the interviewer is an amazing judge of character, or just has read access to your source code. In this situation, stealing might be a great option \emph{as an epsilon-exploration action}; the interviewer may not be able to predict your theft, or may not think punishment makes sense for a one-off anomaly. However, stealing is clearly a bad idea as a normal action, because you will be seen as much less reliable and trustworthy.

\subsection{Viewing the problem from outside}

\noindent \label{dt6}If we do not learn counterfactuals from epsilon-exploration, then, it seems we have no guarantee of learning realistic counterfactuals at all. But if we do learn from epsilon-exploration, it appears we still get things wrong in some cases. Switching to a probabilistic setting doesn't cause the agent to reliably make “reasonable” choices, and neither does forced exploration.

But writing down examples of ``correct'' counterfactual reasoning doesn't seem hard from the outside! 

Maybe that's because from ``outside'' we always have a dualistic perspective. We are in fact sitting outside of the problem, and we've defined it as a function of an agent. However, an agent can't solve the problem in the same way from inside. From its perspective, its functional relationship with the environment isn't an observable fact. This is why counterfactuals are called ``counterfactuals,'' after all. 

 

When we introduced the five-and-ten problem, we first described the problem, and then supplied an agent. When one agent doesn't work well, we could consider a different agent. 

Finding a way to succeed at a decision problem involves finding an agent that when plugged into the problem takes the right action. The fact that we can even consider putting in different agents means that we have already carved the universe into an ``agent'' part, plus the rest of the universe with a hole for the agent---which is most of the work! 

\label{dt7} Are we just fooling ourselves due to the way we set up decision problems, then? Are there no ``correct'' counterfactuals? 

Well, maybe we \emph{are} fooling ourselves. But there is still something we are confused about! ``Counterfactuals are subjective, invented by the agent'' doesn't dissolve the mystery. There is \emph{something} intelligent agents do, in the real world, to make decisions.

So we are not talking about agents that know their own actions because we worry that intelligent machines will run into problems with inferring their own actions in the future. Rather, the possibility of knowing one's own actions illustrates something confusing about determining the consequences of actions---a confusion which shows up even in the very simple case where everything about the world is known and one only needs to choose the larger pile of money.

For all that, \emph{humans} don't seem to run into any trouble taking the \$10. Can we take any inspiration from how humans make decisions?

Well, suppose you are actually asked to choose between \$10 and \$5. You know that you will take the \$10. How do you reason about what \emph{would} happen if you took the \$5 instead?

It seems easy if you can separate yourself from the world, so that you only think of external consequences (getting \$5).

If you think about \emph{yourself} as well, the counterfactual starts seeming a bit more strange or contradictory. Maybe you have some absurd prediction about what the world would be like if you took the \$5---like, "I'd have to be blind!" That's alright, though. In the end, you still see that taking the \$5 would lead to bad consequences, and you still take the \$10, so you're doing fine.

The challenge for formal agents is that an agent can be in a similar position, except it is taking the \$5, knows it is taking the \$5, and can't figure out that it should be taking the \$10 instead, because of the absurd predictions it makes about what happens when it takes the \$10.

It seems hard for a human to end up in a situation like that; yet when we try to write down a formal reasoner, we keep running into this kind of problem. So it indeed seems like human decision-making is doing something here that we don't yet understand.

\subsection{Newcomblike problems}

\noindent If you are an embedded agent, then you should be able to think about yourself, just like you think about other objects in the environment. And other reasoners in your environment should be able to think about you too.

In the five-and-ten problem, we saw how messy things can get when an agent knows its own action before it acts. But this is hard to avoid for an embedded agent.

It's especially hard not to know your own action in standard Bayesian settings, \textcolor{o}{which assume logical omniscience}. A probability distribution assigns probability 1 to any fact which is logically true. So if a Bayesian agent \textcolor{b}{knows its own source code}, then it should know its own action However, realistic agents who are not logically omniscient may run into the same problem. Logical omniscience forces the issue, but rejecting logical omniscience doesn't eliminate the issue.

Epsilon-exploration does seem to solve that problem in many cases, by ensuring that agents have uncertainty about their choices and that the things they expect are based on experience. However, as we saw in the security guard example, even epsilon-exploration seems to steer us wrong when the results of exploring randomly differ from the results of acting reliably.

Examples which go wrong in this way seem to involve another part of the environment that behaves like you—such as another agent very similar to yourself, or a sufficiently good model or simulation of you. These are called \emph{Newcomblike problems}; an example is the Twin Prisoner's Dilemma mentioned above.

If the five-and-ten problem is about cutting a you-shaped piece out of the world so that the world can be treated as a function of your action, Newcomblike problems are about what to do when there are several approximately you-shaped pieces in the world.

One idea is that exact copies should be treated as 100\% under your ``logical control''. For approximate models of you, or merely similar agents, control should drop off sharply as \textcolor{o}{logical correlation} decreases. But how does this work?

Newcomblike problems are difficult for almost the same reason as the self-reference issues discussed so far: prediction. With strategies such as epsilon-exploration, we tried to limit the self-knowledge of the \emph{agent} in an attempt to avoid trouble. But the presence of powerful predictors in the environment reintroduces the trouble. By choosing what information to share, predictors can manipulate the agent and choose their actions for them.

If there is something which can predict you, it might \emph{tell} you its prediction, or related information, in which case it matters what you do \emph{in response} to various things you could find out.

Suppose you decide to do the opposite of whatever you're told. Then it isn't possible for the scenario to be set up in the first place. Either the predictor isn’t accurate after all, or alternatively, the predictor does not share their prediction with you.

On the other hand, suppose there is some situation where you do act as predicted. Then the predictor can control how you'll behave, by controlling what prediction they tell you.

So, on the one hand, a powerful predictor can control you by selecting between the consistent possibilities. On the other hand, you are the one who chooses your pattern of responses in the first place. This means that you can set them up to your best advantage.

\subsection{Observation counterfactuals}

\noindent So far, we have been discussing action counterfactuals---how to anticipate consequences of different actions. This discussion of controlling your responses introduces the \emph{observation counterfactual}---imagining what the world would be like if different facts had been observed.

Even if there is no one telling you a prediction about your future behavior, observation counterfactuals can still play a role in making the right decision. Consider the following game:

Alice receives a card at random which is either High or Low. She may reveal the card if she wishes. Bob then gives his probability $p$ that Alice has a high card. Alice always loses $p^2$ dollars. Bob loses $p^2$ if the card is low, and $(1-p)^2$ if the card is high. 

Bob has a proper scoring rule, so does best by giving his true belief. Alice just wants Bob's belief to be as much toward ``low'' as possible. 

Suppose Alice will play only this one time. She sees a low card. Bob is good at reasoning about Alice, but is in the next room and so can't read any tells. Should Alice reveal her card? 

Since Alice's card is low, if she shows it to Bob, she will lose no money, which is the best possible outcome. However, this means that in the counterfactual world where Alice sees a high card, she wouldn't be able to keep the secret---she might as well show her card in that case too, since her reluctance to show it would be as reliable a sign of ``high''.

On the other hand, if Alice doesn't show her card, she loses 25\textcent---but then she can use the same strategy in the other world, rather than losing \$1. So, before playing the game, Alice would want to visibly commit to not reveal; this makes expected loss 25\textcent, whereas the other strategy has expected loss 50\textcent. By taking observation counterfactuals into account, Alice is able to keep secrets---without them, Bob could perfectly infer her card from her actions. 

This game is equivalent to the decision problem called counterfactual mugging \citep{Nesov:2009:mugging,Garrabrant:2018:poker}.

\textbf{Updateless decision theory} (UDT) is a proposed decision theory which can keep secrets in the high/low card game. UDT does this by recommending that the agent do whatever would have seemed wisest before—whatever your \textcolor{b}{earlier self} would have committed to do.\footnote{See \citet{Dai:2009:udt} and \citet{Yudkowsky:2017:fdt}.}

UDT also performs well in Newcomblike problems. Could something like UDT be related to what humans are doing, if only implicitly, to get good results on decision problems? Or, if it's not, could it still be a good model for thinking about decision-making?

Unfortunately, there are still some pretty deep difficulties here. UDT is an elegant solution to a fairly broad class of decision problems, but it only makes sense if the earlier self can foresee \textcolor{o}{all possible situations}.

This works fine in a Bayesian setting where the prior already contains all possibilities within itself. However, there may be no way to do this in a realistic embedded setting. An agent has to be able to think of \emph{new possibilities}---meaning that its earlier self doesn't know enough to make all the decisions. 

And with that, we find ourselves squarely facing the problem of \textcolor{o}{\emph{embedded world-models}}.

\section{Embedded world-models}\label{sec:ewm}

An agent which is larger than its environment can:

\begin{itemize}
 \tightlist
  \item Hold an exact model of the environment in its head.
  \item Think through the consequences of every potential course of action.
  \item If it doesn't know the environment perfectly, hold every \emph{possible} way the environment could be in its head, as is the case with Bayesian uncertainty.
\end{itemize}

\noindent All of these are typical of notions of rational agency.\footnote{For a general discussion, see \citet{Soares:2015:models} and \citet{Demski:2018:explanation}.}

An embedded agent can't do any of those things, at least not in any straightforward way. 


One difficulty is that, since the agent is part of the environment, modeling the environment in every detail would require the agent to model itself in every detail, which would require the agent's self-model to be as ``big'' as the whole agent. An agent can't fit inside its own head. And the lack of a crisp agent/environment boundary forces us to grapple not only with representing the world at large, but with paradoxes of self-reference.

\textcolor{o}{\textbf{Embedded World-Models}} have to represent the world in a way more appropriate for embedded agents. Problems in this cluster include: 
\begin{itemize}
 \tightlist
  \item the ``realizability'' / ``grain of truth'' problem: the real world isn't in the agent's hypothesis space
  \item logical uncertainty
  \item high-level models
  \item multi-level models
  \item ontological crises
  \item naturalized induction, the problem that the agent must incorporate its model of itself into its world-model
  \item anthropic reasoning, the problem of reasoning with how many copies of yourself exist
\end{itemize}

 \subsection{Realizability}

\label{ewm1} In a Bayesian setting, where an agent's uncertainty is quantified by a probability distribution over possible worlds, a common assumption is ``\textbf{realizability}'': the true underlying environment which is generating the observations is assumed to have at least \emph{some} probability in the prior.

In game theory, this same property is described by saying that a prior has a ``grain of truth'' \citep{Kalai:1993}. It should be noted, however, that there are additional barriers to getting this property in a game-theoretic setting. As such, in their common usage cases, ``grain of truth'' is technically demanding while ``realizability'' is a technical convenience.

Realizability is not totally necessary in order for Bayesian reasoning to make sense. If you think of a set of hypotheses as ``experts'', and the current posterior probability as how much you ``trust'' each expert, then learning according to Bayes' Law,

$$P(h|e) = \frac{P(e|h) \cdot P(h)}{P(e)},$$
ensures a \emph{relative bounded loss} property. 

Specifically, if you use a prior $\pi$, the amount worse you are in comparison to each expert $h$ is at most $\log \pi(h)$, since you assign at least probability $\pi(h) \cdot h(e)$ to seeing a sequence of evidence $e$. Intuitively, $\pi(h)$ is your initial trust in expert $h$, and in each case where it is even a little bit more correct than you, you increase your trust accordingly. The way you do this ensures you assign an expert probability 1 and hence copy it precisely before you lose more than $\log \pi(h)$ compared to it. 

The prior AIXI is based on is the \emph{Solomonoff prior}. It is defined as the output of a universal Turing machine (UTM) whose inputs are coin-flips. In other words, feed a UTM a random program. Normally, we would think of a UTM as only being able to simulate deterministic machines. Here, however, the initial inputs can instruct the UTM to use the rest of the infinite input tape as a source of randomness to simulate a \emph{stochastic} Turing machine. 

Combining this with the previous idea about viewing Bayesian learning as a way of allocating ``trust'' to ``experts'' which meets a bounded loss condition, we can see the Solomonoff prior as a kind of ideal machine learning algorithm which can learn to act like any algorithm you might come up with, no matter how clever. 

For this reason, we shouldn't \emph{necessarily} think of AIXI as ``assuming the world is computable'', even though it reasons via a prior over computations. AIXI achieves bounded loss on its predictive accuracy \emph{as compared with} any computable predictor. We should rather say that AIXI assumes all possible algorithms are computable, not that the world is.

However, lacking realizability can cause trouble if you are looking for anything more than bounded-loss predictive accuracy: 

\begin{itemize}
 \tightlist
  \item the posterior can oscillate forever;
  \item probabilities may not be calibrated;
  \item estimates of statistics such as the mean may be arbitrarily bad;
  \item estimates of latent variables may be bad;
  \item and the identification of causal structure may not work.
\end{itemize}

\noindent So does AIXI perform well without a realizability assumption? We don't know. Despite getting bounded loss for \emph{predictions} without realizability, existing optimality results for its \emph{actions} require an added realizability assumption.

First, if the environment really \emph{is} sampled from the Solomonoff distribution, AIXI gets the \textcolor{r}{maximum expected reward}. But this is fairly trivial; it is essentially the definition of AIXI. Second, if we modify AIXI to take somewhat randomized actions---Thompson sampling---there is an \textcolor{r}{\emph{asymptotic} optimality result} for environments which act like any stochastic Turing machine. In both cases, realizability is assumed in order to prove anything. For details, see \citet{Leike:2016:nonparametric}.

But the point we are raising is \emph{not} ``the world might be uncomputable, so we don't know if AIXI will do well''; this is more of an illustrative case. The concern is rather that AIXI is only able to define intelligence or rationality by constructing an agent \textcolor{o}{\emph{much, much bigger}} than the environment which it has to learn about and act within.


\citet{Orseau:2012:spacetime} provide a way of thinking about this in ``Space-Time Embedded Intelligence''. However, their approach defines the intelligence of an agent in terms of a sort of super-intelligent designer who thinks about reality from outside, selecting an agent to \textcolor{r}{place into the environment}.

Embedded agents don't have the luxury of stepping outside of the universe to think about how to think. What we would like would be a theory of rational belief for \emph{situated} agents which provides foundations that are similarly as strong as the foundations Bayesianism provides for dualistic agents.

As a clarifying point: Imagine a computer science theory person who is having a disagreement with a programmer. The theory person is making use of an abstract model. The programmer is complaining that the abstract model isn’t something you would ever run, because it is computationally intractable. The theory person responds that the point isn’t to ever run it. Instead, the point is to understand some phenomenon which will also be relevant to more tractable things which we would want to run.

We bring this up in order to emphasize that our perspective is much more like the theory person’s, as described. We are not talking about AIXI in order to say ``AIXI is an idealization you can’t run''. The answers to the puzzles we're pointing at don’t need to run; we just want to understand this set of phenomena. However, sometimes a factor that makes some theoretical models less tractable also makes a model too dissimilar from the phenomenon we are interested in.

The qualitative way AIXI wins games is by assuming we can do true Bayesian updating over a hypothesis space, assuming the world is in our hypothesis space, and so on. As such, AIXI can tell us something about the aspect of realistic agency that is approximately doing Bayesian updating over an approximately-good-enough hypothesis space. But embedded agents don’t just need approximate solutions to that problem; they need to solve several problems that are different in kind from that problem.

\subsection{Self-reference}

One major obstacle a theory of embedded agency must deal with is \textbf{self-reference}. Paradoxes of self-reference such as the liar paradox make it not just wildly impractical, but in a certain sense impossible for an agent’s world-model to accurately reflect the world.

The liar paradox concerns the status of the sentence ``This sentence is not true''. If it were true, it must be false; and if not true, it must be true.

The difficulty comes in part from trying to draw a map of a territory which includes the map itself. This is fine if the world ``holds still'' for us; but because the map is in the world, \textcolor{r}{different maps create different worlds}.

Suppose our goal is to make an accurate map of the final route of a road which is currently under construction. Suppose we also know that the construction team will get to see our map, and that construction will proceed so as to disprove whatever map we make. This puts us in a liar-paradox-like situation.

Problems of this kind become relevant for \textcolor{r}{decision-making} in the theory of games. A simple game of rock-paper-scissors can introduce a liar paradox if the players are trying to win, and can predict each other better than chance.

Game theory solves this type of problem with game-theoretic equilibria. But the problem ends up returning in a different way.

We noted that the problem of realizability takes on a different character in the context of game theory. In an ML setting, realizability is a potentially \emph{unrealistic} assumption, but can usually be assumed consistently nonetheless.

In game theory, on the other hand, the assumption itself may be inconsistent. This is because games commonly yield paradoxes of self-reference.

Because there are so many agents, it is no longer possible in game theory to conveniently make an ``agent'' a thing which is larger than a world. Game theorists are therefore forced to investigate notions of rational agency which can handle a large world.

Unfortunately, this is done by splitting up the world into ``agent'' parts and ``non-agent'' parts, and handling the agents in a special way. This is almost as bad as dualistic models of agency.

In rock-paper-scissors, the liar paradox is resolved by stipulating that each player play each move with $1/3$ probability. If one player plays this way, then the other loses nothing by doing so. This way of introducing probabilistic play to resolve would-be paradoxes of game theory is called a \emph{Nash equilibrium}.

We can use Nash equilibria to prevent the assumption that the agents correctly understand the world they’re in from being inconsistent. However, this works just by telling the agents what the world looks like. What if we want to model agents who learn about the world, more like AIXI?

The \textbf{grain of truth problem} is the problem of formulating a reasonably bound prior probability distribution which would allow agents playing games to place \emph{some} positive probability on each other’s true (probabilistic) behavior, without knowing it precisely from the start.

Until recently, known solutions to the problem were quite limited. Fallenstein, Taylor, and Christiano's (\citeyear{Fallenstein:2015:oraclesai}) ``Reflective Oracles'' provide a very general solution. For details, see \citet{Leike:2016:grain}.

It may seem as though stochastic Turing machines are perfectly fine tools for representing Nash equilibria. However, if you're trying to produce Nash equilibria as a result of reasoning about other agents, you'll run into trouble. If each agent models the other's computation and tries to run it to see what the other agent does, the result is an infinite loop.

There are some questions Turing machines just can't answer---in particular, questions about the behavior of Turing machines. The halting problem is the classic example.

Turing studied ``oracle machines'' to examine what would happen if we could answer such questions. An oracle is like a book containing some answers to questions which we were unable to answer before. But ordinarily, we get a hierarchy. Type B machines can answer questions about whether type A machines halt, type C machines have the answers about types A and B, and so on, but no machines have answers about their own type.\citep{Shore:1999}

Reflective oracles work by twisting the ordinary Turing universe back on itself, so that rather than an infinite hierarchy of ever-stronger oracles, you define an oracle that serves as its own oracle machine. This would normally introduce contradictions, but reflective oracles avoid this by randomizing their output in cases where they would run into paradoxes. So reflective oracle machines are stochastic, but they are more powerful than regular stochastic Turing machines.

This is how reflective oracles address the problems we mentioned earlier of a map that's itself part of the territory: randomize. Reflective oracles also solve the problem with game-theoretic notions of rationality we mentioned earlier. The reflective oracles framework allows agents to be reasoned about in the same manner as other parts of the environment, rather than treating agents as a fundamentally special case. All agents can just be modeled as computations-with-oracle-access.

However, models of rational agents based on reflective oracles still have several major limitations. One of these is that agents are required to have unlimited processing power, just like AIXI, and so are assumed to know all of the consequences of their own beliefs.

In fact, knowing all of the consequences of one's beliefs---a property known as \emph{logical omniscience}---turns out to be rather core to classical Bayesian rationality.

\subsection{Logical uncertainty}

\label{ewm2} So far, we've been talking in a fairly naive way about the agent having beliefs about hypotheses, and the real world being or not being in the hypothesis space. It isn't really clear what any of this means. 

Depending on how we define the relevant terms, it may actually be quite possible for an agent to be smaller than the world and yet contain the right world-model---it might know the true physics and initial conditions, but only be capable of inferring their consequences very approximately.

Humans are certainly used to living with shorthands and approximations. But realistic as this scenario may be, it is not in line with what it usually means for a Bayesian to know something. A Bayesian knows the consequences of all of its beliefs.

Uncertainty about the consequences of your beliefs is \textbf{logical uncertainty} \citep{Soares:2015:questions}. In this case, the agent might be empirically certain of a unique mathematical description pinpointing which universe she is in, while being logically uncertain of most of the consequences of that description. 

Modeling logical uncertainty requires us to have a combined theory of logic (reasoning about implications) and probability (degrees of belief).

Logic and probability theory are two great triumphs in the codification of rational thought. Logic provides the best tools for thinking about \textcolor{b}{self-reference}, while probability provides the best tools for thinking about \textcolor{r}{decision-making}. However, the two don't work together as well as one might think.

\vspace{2mm}

\begin{center}
\includegraphics[width=0.9\textwidth]{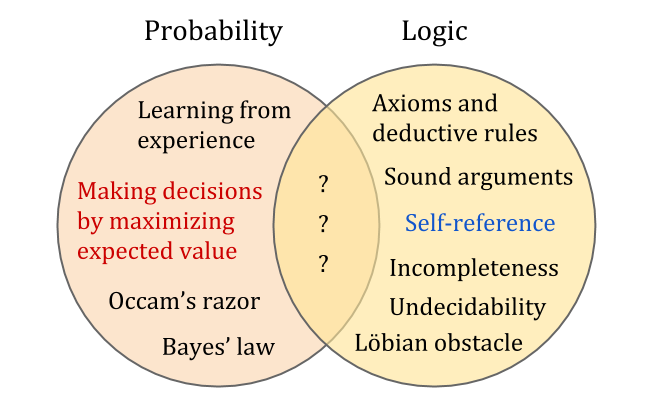}
\end{center}

\vspace{2mm}

The two may seem superficially compatible, since probability theory is an extension of Boolean logic. However, G\"{o}del's first incompleteness theorem shows that any sufficiently rich logical system is incomplete: not only does it fail to decide every sentence as true or false, but it also has no computable extension which manages to do so.\footnote{For further discussion of the problems this creates for probability theory, see \citet{Demski:2018:untrollable}.}

This also applies to probability distributions: no computable distribution can assign probabilities in a way that's consistent with a sufficiently rich theory. This forces us to choose between using an uncomputable distribution, or using a distribution which is inconsistent.

This may seem like an easy choice: the inconsistent theory is at least computable, and we are after all trying to develop a theory of logical \emph{non}-omniscience. We can just continue to update on facts which we prove, bringing us closer and closer to consistency.

Unfortunately, this doesn't work out so well, for reasons which connect back to realizability. Recall that there are \emph{no} computable probability distributions consistent with all consequences of sound theories. So our non-omniscient prior doesn't even contain a single correct hypothesis. This causes pathological behavior as we condition on more and more true mathematical beliefs. Beliefs wildly oscillate rather than approaching reasonable estimates.

Taking a Bayesian prior on mathematics, and updating on whatever we prove, does not seem to capture mathematical intuition and heuristic conjecture very well---unless we restrict the domain and craft a sensible prior.

Probability is like a scale, with worlds as weights. An observation eliminates some of the possible worlds, removing weights and shifting the balance of beliefs. 

Logic is like a tree, growing from the seed of axioms according to inference rules. For real-world agents, the process of growth is never complete; you never know all the consequences of each belief.

Without knowing how to combine the two, we can't characterize reasoning probabilistically about math. But the ``scale versus tree'' problem also means that we don’t know how ordinary empirical reasoning works.

Bayesian hypothesis testing requires each hypothesis to clearly declare which probabilities it assigns to which observations. This allows us to determine how much to rescale the odds when we make an observation. If we don’t know the consequences of a belief, we don’t know how much credit to give the belief for making predictions.

This is like not knowing where to place the weights on the scales of probability. We could try putting weights on both sides of the scales until a proof rules one out, but then the beliefs just oscillate forever rather than doing anything useful.

This forces us to grapple directly with the problem of a world that is larger than the agent. We want some notion of boundedly rational beliefs about uncertain consequences; but \emph{any} computable beliefs about logic must have left out \emph{something}, since the tree of logical implications will grow larger than any container.

For a Bayesian, the scales of probability are balanced in precisely such a way that no Dutch book can be made against them—no sequence of bets that are a sure loss \citep{Yudkowsky:2017:coherent}. But one can only account for all Dutch books if one knows all of the consequences of one's beliefs. Absent that, someone who has explored other parts of the tree can Dutch-book you.

But human mathematicians don’t seem to run into any special difficulty in reasoning about mathematical uncertainty, any more than we do with empirical uncertainty. So what characterizes good reasoning under mathematical uncertainty, if not immunity to making bad bets?

One answer is to weaken the notion of Dutch books so that we only allow bets based on quickly computable parts of the tree. This is one of the ideas behind Garrabrant et al.’s 
(\citeyear{Garrabrant:2016:li}) ``Logical Induction'', an early attempt at developing an analog for Solomonoff induction that incorporates mathematical uncertainty.

\subsection{High-level models}

\label{ewm3} Another consequence of the fact that the world is bigger than you is that you need to be able to use \textbf{high-level world models}: models which involve things like tables and chairs \citep{Yudkowsky:2015:ontology}.

This is related to the classical symbol grounding problem; but since we want a formal analysis which increases our \textcolor{b}{trust} in some system, the kind of model which interests us is somewhat different. This also relates to \textcolor{g}{transparency} and \textcolor{b}{informed oversight} \citep{Christiano:2016:oversight}: world-models should be made out of understandable parts. 

A related question is how high-level reasoning and low-level reasoning relate to each other and to intermediate levels: \textbf{multi-level world models}. 

Standard probabilistic reasoning does not provide a very good account of this topic. It's as though we have different Bayes nets which describe the world at different levels of accuracy, and processing power limitations force us to primarily use the less accurate ones, so that we must decide how to jump to the more accurate as needed.

Additionally, the models at different levels don't line up perfectly, so we have a problem of translating between them; and the models may have serious contradictions between them. This may be fine, since high-level models are understood to be approximations anyway, or it could signal a serious problem in the higher- or lower-level models, requiring their revision. 

This is especially interesting in the case of \textbf{ontological crises} \citep{de-Blanc:2011:ontological}, in which objects we value turn out not to be a part of ``better'' models of the world. 

It seems fair to say that everything humans value exists in high-level models only, which from a reductionistic perspective are ``less real'' than atoms and quarks. However, \emph{because} our values are not defined on the low level, we are able to keep our values even when our knowledge of the low level radically shifts.\footnote{We would also like to be able to say something about what happens to values if the \emph{high} level radically shifts.}

\label{ewm4} Another critical aspect of embedded world-models is that the agent itself must be in the model, since the agent seeks to understand the world, and the world cannot be fully separated from oneself. This opens the door to difficult problems of \textcolor{b}{self-reference} and \textcolor{r}{anthropic decision theory}.

\textbf{Naturalized induction} is the problem of learning world-models which include yourself in the environment. This is challenging because \mkbibparens{as Caspar Oesterheld [\citeyear{Osterheld:2017:naturalized}] has put it} there is a type mismatch between ``mental stuff'' and ``physics stuff''. 

AIXI conceives of the environment as if it were made with \textcolor{r}{a slot which the agent fits into}. We might intuitively reason in this way, but we can also understand a physical perspective from which this looks like a bad model. We might imagine instead that the agent separately represents: \textcolor{b}{self-knowledge} available to introspection; hypotheses about what the universe is like; and a ``bridging hypothesis'' connecting the two \citep{Bensinger:2013:phenom}.

There are interesting questions of how this could work. There's also the question of whether this is the right structure at all. It's certainly not how we imagine babies learning. 

Thomas \citet{Nagel:1986:nowhere} would say that this way of approaching the problem involves ``views from nowhere''; each hypothesis posits a world as if seen from outside. This is perhaps a strange thing to do. 

\vspace{4mm}

\label{ewm5} A special case of agents needing to reason about themselves is agents needing to reason about their \textcolor{b}{\emph{future} self}.

To make long-term plans, agents need to be able to model how they'll act in the future, and have a certain kind of \textcolor{b}{\emph{trust}} in their future goals and reasoning abilities. This includes trusting future selves that have learned and grown a great deal. 

In a traditional Bayesian framework, ``learning'' means Bayesian updating. But as we noted, Bayesian updating requires that the agent \emph{start out} large enough to fully model many different ways the world might be, and learn by ruling some of these models out. Embedded agents need resource-limited, logically uncertain updates, which don't work like this. 

Unfortunately, Bayesian updating is the main way we know how to think about an agent progressing through time as one unified agent. The Dutch book justification for Bayesian reasoning is effectively saying that this kind of updating is the only way to not have the agent's actions on Monday work at cross purposes, at least a little, to the agent's actions on Tuesday.

Embedded agents are non-Bayesian, and non-Bayesian agents tend to get into wars with their future selves. This brings us to our next set of problems: \textcolor{b}{\emph{robust delegation}}.

\section{Robust delegation}\label{sec:rd}

Because \textcolor{o}{the world is big}, the agent as it is may be inadequate to accomplish its goals, including in its ability to think. 

Because the agent is \textcolor{g}{made of parts}, it can improve itself and become more capable. 

Improvements can take many forms: The agent can make tools, the agent can make successor agents, or the agent can just learn and grow over time. However, the successors or tools need to be more capable for this to be worthwhile. 

This gives rise to a special type of principal/agent problem: You have an initial agent, and a successor agent. The initial agent gets to decide exactly what the successor agent looks like. The successor agent, however, is much more intelligent and powerful than the initial agent. We want to know how to have the successor agent robustly optimize the initial agent's goals.

We will consider three different forms this principal/agent problem can take:
\begin{itemize}
 \tightlist
  \item In the \emph{AI alignment problem} of \citet{Soares:2015:align,Bostrom:2014:superintelligence}, a human is trying to build an AI system which can be trusted to help with the human's goals.
  \item In the \emph{tiling agents problem} of \citet{Yudkowsky:2013:tiling}, an agent is trying to make sure it can trust its future selves to help with its own goals.
  \item Or we can consider a harder version of the tiling problem---\emph{reflective stability of goal systems under self-improvement}---where an AI system has to build a successor which is more intelligent than itself, while still being trustworthy and helpful.
\end{itemize}
For a human analogy which involves no AI, we can consider the problem of succession in royalty, or more generally the problem of setting up organizations to achieve desired goals without losing sight of their purpose over time. The difficulty seems to be twofold:

First, a human or AI agent may not fully understand itself and its own goals. If an agent cannot write out what it wants in exact detail, then this makes it difficult to guarantee that its successor will robustly help with the goal.

Second, the idea behind delegating work is that you not have to do all the work yourself. You want the successor to be able to act with some degree of autonomy, including learning new things that you don't know, and wielding new skills and capabilities. In the limit, an excellent formal account of robust delegation should be able to handle arbitrarily capable successors without throwing up any errors---such as a human or an AI system building an extraordinarily intelligent AI system, or such as an agent that just continues learning and growing for so many years that it ends up much smarter than its past self.

The problem is not (just) that the successor agent might be malicious. The problem is that we don't even know what it means not to be. 

This problem seems hard from both points of view. 
The initial agent needs to figure out how reliable and trustworthy something more powerful than it is, which seems very hard. But the successor agent has to figure out what to do in situations that the initial agent can't even understand, and try to respect the goals of something that the successor can see is inconsistent, which also seems very hard \citep{Armstrong:2017:occam}.

At first, this may look like a less fundamental problem than ``\textcolor{r}{make decisions}'' or ``\textcolor{o}{have models}''. But the view on which there are multiple forms of the ``build a successor'' problem is itself a dualistic view. To an embedded agent, the future self is not privileged; it is just another part of the environment. There is no deep difference between building a successor that shares your goals, and just making sure your own goals stay the same over time.

For this reason, although we talk about ``initial'' and ``successor'' agents, remember that this is not just about the narrow problem humans currently face of aiming a successor. This is about the fundamental problem of being an agent that persists and learns over time.

We call this cluster of problems \textcolor{b}{\textbf{Robust Delegation}}. Examples include:
\begin{itemize}
 \tightlist
  \item Vingean reflection
  \item the tiling problem
  \item averting Goodhart's law
  \item value loading
  \item corrigibility
  \item informed oversight
\end{itemize}

\subsection{Vingean reflection}

\label{rd1} Imagine that you are playing Hadfield-Menell et al.'s (\citeyear{Hadfield-Menell:2016:cirl}) Cooperative Inverse Reinforcement Learning (CIRL) game  with a toddler.

The idea behind CIRL is to define what it means for a smart learning system to collaborate with a human. The AI system tries to pick helpful actions, while simultaneously trying to figure out what the human wants. 

A lot of current work on robust delegation comes from the goal of aligning AI systems with what humans want, so we usually think about this from the point of view of the human. But now consider the problem faced by a smart AI system, where they’re trying to help someone who is very confused about the universe. Imagine trying to help a toddler optimize their goals.

\begin{itemize}
 \tightlist
  \item From your standpoint, the toddler may be too irrational to be seen as optimizing anything.
  \item The toddler may have an ontology in which it is optimizing something, but you can see that ontology doesn't make sense.
  \item Maybe you notice that if you set up questions in the right way, you can make the toddler seem to want almost anything.
\end{itemize}

\noindent Part of the problem is that the ``helping'' agent has to be \textcolor{o}{\emph{bigger}} in some sense in order to be more capable; but this seems to imply that the ``helped'' agent can't be a very good supervisor for the ``helper''. 

For example, \textcolor{r}{updateless decision theory} eliminates dynamic inconsistencies in decision theory by, rather than maximizing expected utility of your action \emph{given} what you know, maximizing expected utility \emph{of reactions} to observations, from a state of ignorance. 

Appealing as this may be as a way to achieve reflective consistency, it creates a strange situation in terms of computational complexity: If \textcolor{r}{actions} are type $A$, and \textcolor{r}{observations} are type $O$, reactions to observations are type $O \to A$---a much larger space to optimize over than $A$ alone. And we're expecting our \textcolor{o}{\emph{smaller}} self to be able to do that! This seems like a major problem.\footnote{See \citet{Garrabrant:2017:updateless} for further discussion.}

One way to more crisply state the problem is: We should be able to trust that our future self is applying its intelligence to the pursuit of our goals \emph{without} being able to predict precisely what our future self will do. This criterion is called \textbf{Vingean reflection} \citep{Fallenstein:2015:vingean}. 

For example, you might plan your driving route before visiting a new city, but you do not plan your steps. You plan to some level of detail, and trust that your future self can figure out the rest. 

Vingean reflection is difficult to examine via classical Bayesian decision theory because Bayesian decision theory assumes \textcolor{o}{logical omniscience}. Given logical omniscience, the assumption ``the agent knows its future actions are rational'' is synonymous with the assumption ``the agent knows its future self will act according to one particular optimal policy which the agent can predict in advance''. 

We have some limited models of Vingean reflection \mkbibparens{see \citet{Yudkowsky:2013:tiling}}. A successful approach must walk the narrow line between two problems: 

\begin{itemize}
 \tightlist
  \item \emph{The L\"{o}bian Obstacle}: Agents who trust their future self because they trust the output of their own reasoning are inconsistent.
  \item \emph{The Procrastination Paradox}: Agents who trust their future selves \emph{without} reason tend to be consistent but unsound and untrustworthy, and will put off tasks forever because they can do them later.
\end{itemize}

\noindent The Vingean reflection results so far apply only to limited sorts of decision procedures, such as satisficers aiming for a threshold of acceptability. So there is plenty of room for improvement, getting tiling results for more useful decision procedures and under weaker assumptions. 

\label{rd2} However, there is more to the robust delegation problem than just tiling and Vingean reflection. When you construct another agent, rather than delegating to your future self, you more directly face a problem of \textbf{value loading} \citep{Soares:2015:value}.

\subsection{Goodhart's law}

The main problems in the context of value loading are: \begin{itemize}
 \tightlist
  \item We don't know what we want \citep{Yudkowsky:2015:cov}.
  \item Optimization amplifies slight differences  between what we say we want and what we really want \citep{Garrabrant:2018:amplifies}.
\end{itemize}

\noindent The misspecification-amplifying effect is known as \textbf{Goodhart's law}, named for Charles Goodhart's (\citeyear{Goodhart:1975:monetary}) observation: ``Any observed statistical regularity will tend to collapse once pressure is placed upon it for control purposes.'' 

When we specify a target for optimization, it is reasonable to expect it to be correlated with what we want---highly correlated, in some cases. Unfortunately, however, this does not mean that optimizing it will get us closer to what we want, especially at high levels of optimization. 

\label{rd3} As described by \citet{Manheim:2018:goodhart}, there are (at least) four mechanisms behind Goodhart's law: regressional, extremal, causal, and adversarial. 

\vspace{3mm}

\begin{center}
  \includegraphics[width=0.9\textwidth]{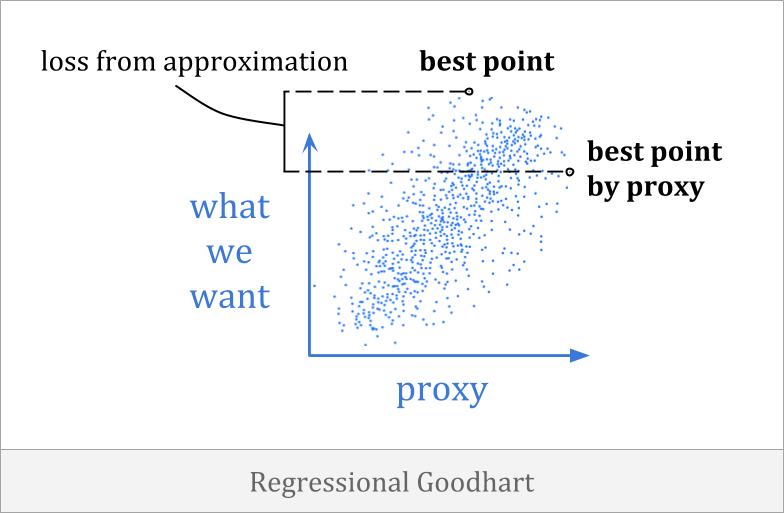}
\end{center}

\vspace{3mm}

\noindent \emph{Regressional Goodhart} occurs when there is a less than perfect correlation between the proxy and the goal. It is more commonly known as the \emph{optimizer's curse} \citep{Smith:2006:curse}, and it is related to regression to the mean.

An example of regressional Goodhart is that you might draft players for a basketball team based on height alone. This isn't a perfect heuristic, but there is a correlation between height and basketball ability, which you can make use of in making your choices.

It turns out that, in a certain sense, you will be predictably disappointed if you expect the general trend to hold up as strongly for your selected team. Stated in statistical terms: An unbiased estimate of $y$ given $x$ is not an unbiased estimate of $y$ when we select for the best $x$. In that sense, we can expect to be disappointed when we use $x$ as a proxy for $y$ for optimization purposes.
 
\vspace{3mm}
 
\begin{center}
  \includegraphics[width=0.9\textwidth, cfbox=gray]{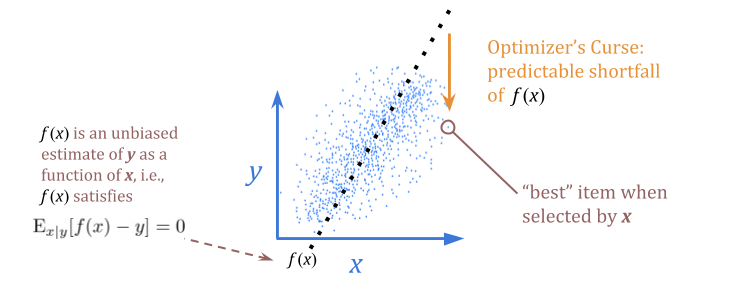}
\end{center}

\vspace{3mm}

\noindent Using a Bayes estimate instead of an unbiased estimate, we can eliminate this sort of predictable disappointment. The Bayes estimate accounts for the noise in $x$, bending toward typical $y$ values.

\vspace{3mm}

\begin{center}
  \includegraphics[width=0.9\textwidth, cfbox=gray]{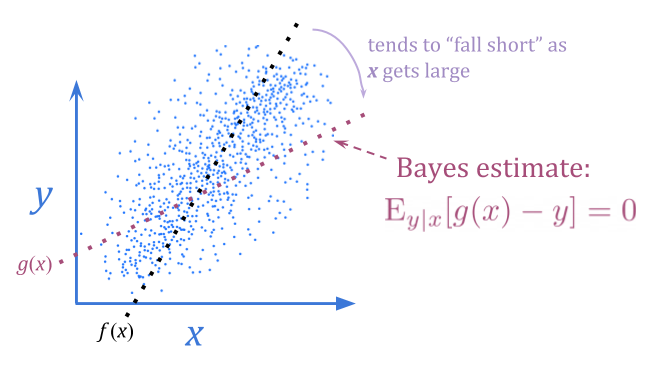}
\end{center}

\vspace{3mm}

\noindent This does not necessarily allow us to achieve a better $y$ value, since we still only have the information content of $x$ to work with. However, it sometimes may. If $y$ is normally distributed with variance $1$, and $x$ is $y \pm 10$ with even odds of $+$ or $-$, a Bayes estimate will give better optimization results by almost entirely removing the noise.

Regressional Goodhart seems like the easiest form of Goodhart to beat---just use Bayes! However, there are two problems with this solution:\begin{itemize}
 \tightlist
  \item Bayesian estimators are very often intractable in cases of interest.
  \item It only makes sense to trust the Bayes estimate under a \textcolor{o}{realizability} assumption.
\end{itemize}
A case where both of these problems become critical is computational learning theory. It often isn't computationally feasible to calculate the Bayesian expected generalization error of a hypothesis. And even if you could, you would still need to wonder whether your chosen prior reflected the world well enough.

\vspace{3mm}
 
\begin{center}
  \includegraphics[width=0.9\textwidth]{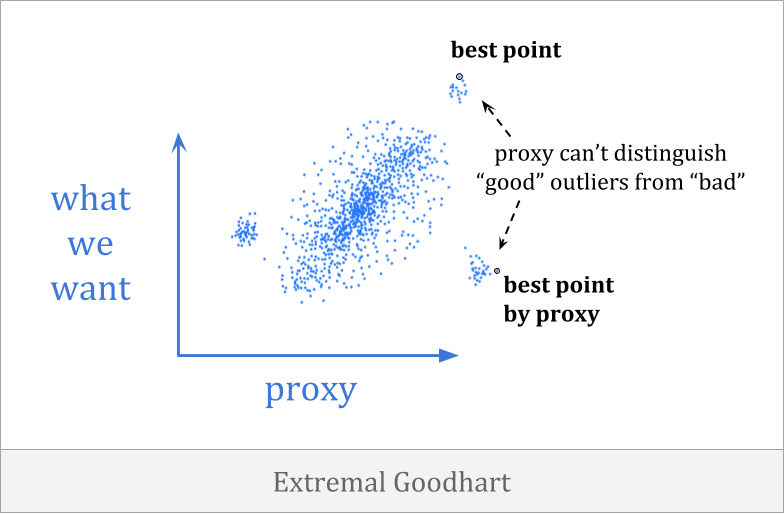}
\end{center}

\vspace{3mm}

\noindent In \emph{extremal Goodhart}, optimization pushes you outside the range where the correlation exists, into portions of the distribution which behave very differently. This manifestation of Goodhart's law is especially scary because it tends to involves optimizers behaving in sharply different ways in different contexts, often with little or no warning. You may not be able to observe the proxy breaking down at all when you have weak optimization, but once the optimization becomes strong enough, you can enter a very different domain.

The difference between extremal Goodhart and regressional Goodhart is related to the classical interpolation/extrapolation distinction. Because extremal Goodhart involves a sharp change in behavior as the system is scaled up, it is harder to anticipate than regressional Goodhart.

As in the regressional case, a Bayesian solution addresses this concern in principle, if you trust a probability distribution to reflect the possible risks sufficiently well. However, the realizability concern seems even more prominent here.

Can a prior be trusted to anticipate problems with proposals, when those proposals have been highly optimized to look good to that specific prior? Certainly a human’s judgment couldn’t be trusted under such conditions---an observation which suggests that this problem will remain even if a system's judgments about values perfectly reflect a human's.

We might say that the problem is this: ``typical'' outputs avoid extremal Goodhart, but ``optimizing too hard'' takes you out of the realm of the typical. But how can we formalize ``optimizing too hard'' in decision-theoretic terms?

\emph{Quantilization} offers a formalization of ``optimize this some, but don't optimize too much'' \citep{Taylor:2015:quantilizers}. Imagine a proxy $V(x)$ as a ``corrupted'' version of the function we really want, $U(x)$. There might be different regions where the corruption is better or worse. Suppose that we can additionally specify a ``trusted'' probability distribution $P(x)$, for which we are confident that the average error is below some threshold $c$. By stipulating $P$ and $c$, we give information about where to find low-error points, without needing to have any estimates of $U$ or of the actual error at any one point. When we select actions from $P$ at random, we can be sure regardless that there’s a low probability of high error.

How do we use this to optimize? A quantilizer selects from $P$, but discarding all but the top fraction $f$; for example, the top 1\%. By quantilizing, we can guarantee that if we overestimate how good something is, we’re overestimating by at most $\frac{c}{f}$ in expectation. This is because in the worst case, all of the overestimation was of the $f$ best options. We can therefore choose an acceptable risk level, $r = \frac{c}{f}$, and set the parameter $f$ as $\frac{c}{r}$.

\vspace{3mm}

\begin{center}
  \includegraphics[width=0.75\textwidth, cfbox=gray]{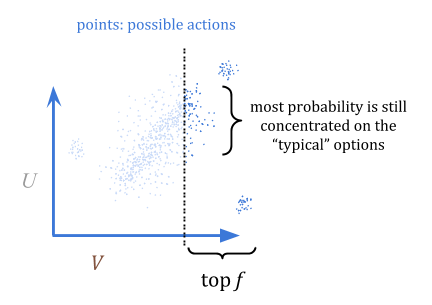}
\end{center}

\vspace{3mm}

\noindent Quantilization is in some ways very appealing, since it allows us to specify safe classes of actions without trusting every individual action in the class---or without trusting \emph{any} individual action in the class. If you have a sufficiently large heap of apples, and there is only one rotten apple in the heap, choosing randomly is still very likely safe. By ``optimizing less hard'' and picking a random good-enough action, we make the genuinely extreme options low-probability. In contrast, if we had optimized as hard as possible, we might have ended up selecting from only bad apples.

However, this approach also leaves a lot to be desired. Where do ``trusted'' distributions come from? How do you estimate the expected error $c$, or select the acceptable risk level $r$? Quantilization is a risky approach because $r$ gives you a knob to turn that will seemingly improve performance, while increasing risk, until (possibly sudden) failure.

Additionally, quantilization doesn't seem likely to \emph{tile}. That is, a quantilizing agent has no special reason to preserve the quantilization algorithm when it makes improvements to itself or builds new agents.

So there seems to be room for improvement in how we handle extremal Goodhart.


\begin{center}
  \includegraphics[width=0.8\textwidth]{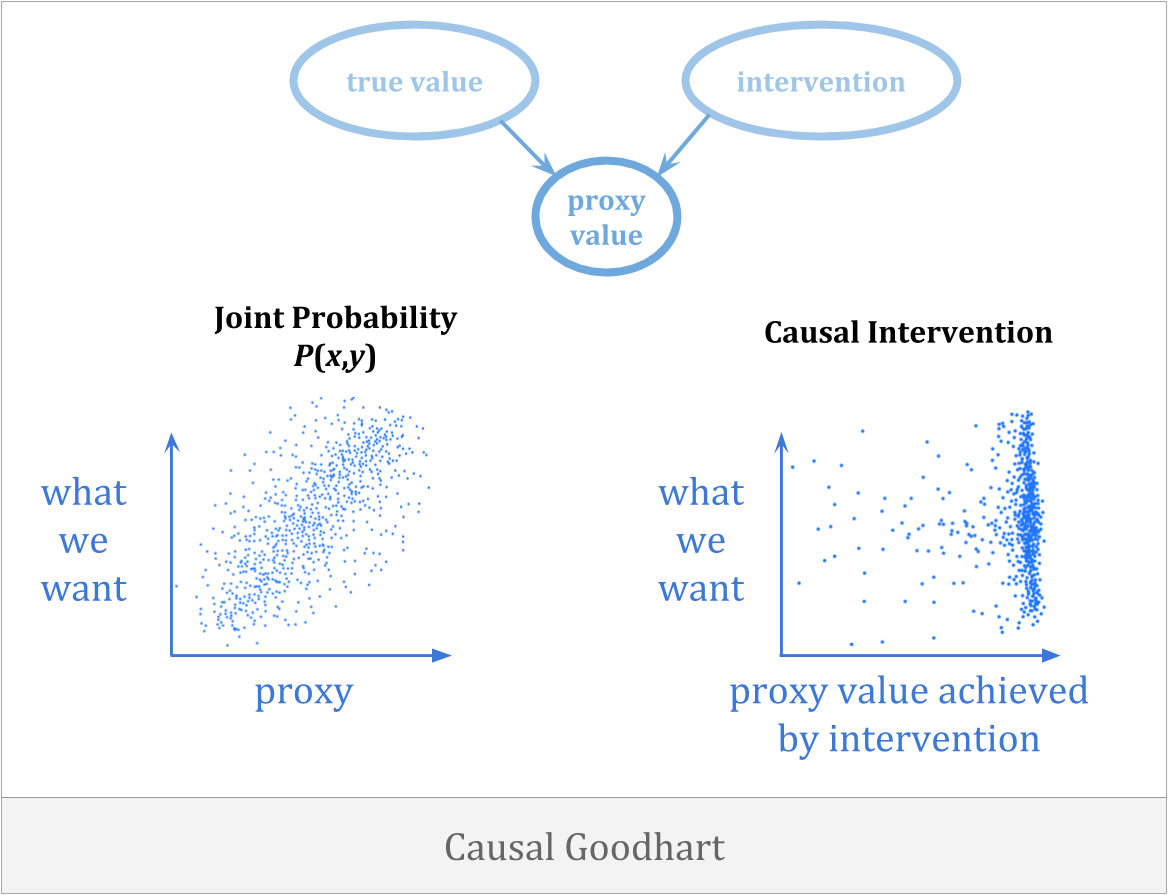}
\end{center}


\noindent Another way optimization can go wrong is when the act of selecting for a proxy breaks the connection to what we care about. \emph{Causal Goodhart} occurs when you observes a correlation between proxy and goal, but upon intervening to increase the proxy, you fail to increase the goal because the observed correlation was not causal in the right way.

An example of causal Goodhart is that you might try to make it rain by carrying an umbrella around. The only way to avoid this sort of mistake is to get \textcolor{r}{counterfactuals}

This might seem like punting to decision theory, but the connection here enriches robust delegation and decision theory alike. Counterfactuals have to address concerns of trust due to tiling concerns---the need for decision-makers to reason about their own future decisions. At the same time, \textcolor{b}{trust} has to address counterfactual concerns because of causal Goodhart.

Once again, one of the big challenges here is \textcolor{o}{realizability}. As we noted in our discussion of embedded world-models, even if you have the right theory of how counterfactuals work in general, Bayesian learning doesn't provide much of a guarantee that you'll learn to select actions well, unless we assume realizability.

\vspace{3mm}

\begin{center}
  \includegraphics[width=0.9\textwidth]{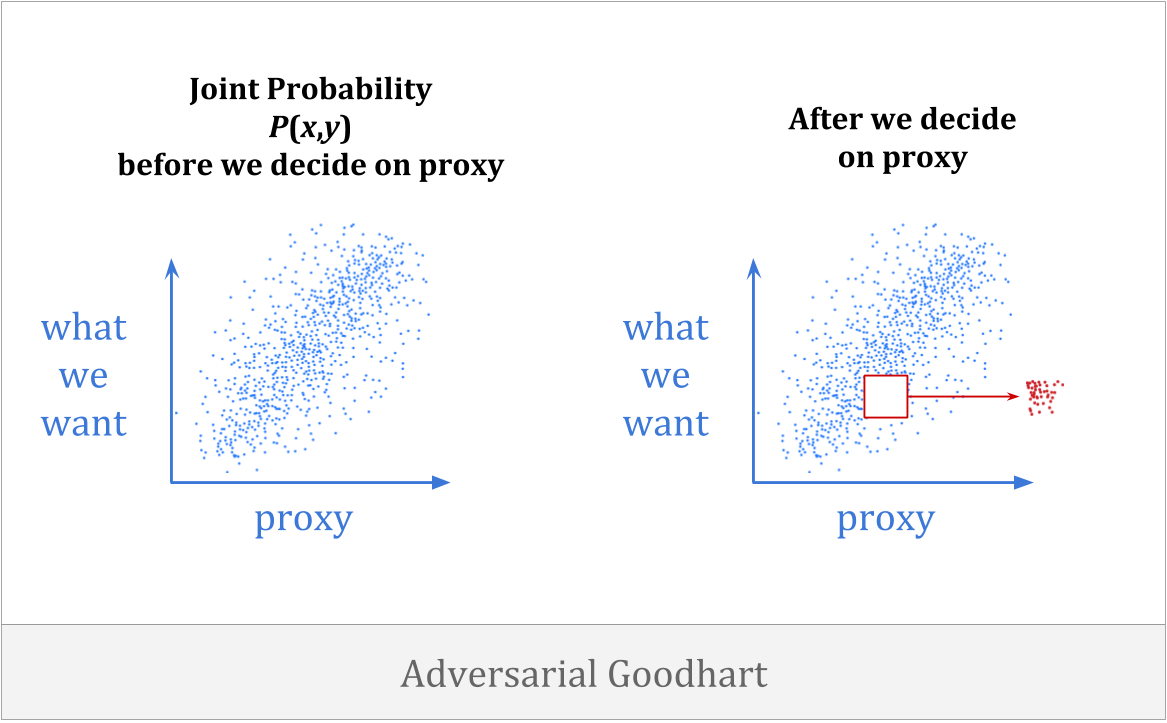}
\end{center}

\vspace{3mm}

\noindent Finally, there is \emph{adversarial Goodhart}, in which agents actively make our proxy worse by intelligently manipulating it.

This category is what people most often have in mind when they interpret Goodhart's remark, and at first glance, it may not seem as relevant to our concerns here. We want to understand in formal terms how agents can trust their future selves, or trust helpers they built from scratch. What does that have to do with adversaries?

The short answer is: when searching in a \textcolor{o}{large} space which is sufficiently rich, there are bound to be some elements of that space which implement adversarial strategies. Understanding optimization in general requires us to understand how sufficiently smart optimizers can avoid adversarial Goodhart. (We'll come back to this point in our discussion of \textcolor{g}{subsystem alignment}.)

The adversarial variant of Goodhart's law is even harder to observe at low levels of optimization, both because the adversaries won’t want to start manipulating until after test time is over, and because adversaries that come from the system’s own optimization won’t show up until the optimization is powerful enough.

These four forms of Goodhart's law work in very different ways---and roughly speaking, they tend to start appearing at successively higher levels of optimization power, beginning with regressional Goodhart and proceeding to causal, then extremal, then adversarial. So be careful not to think you have conquered Goodhart’s law because you have solved some of them.

\subsection{Stable pointers to value}

\label{rd4} Besides anti-Goodhart measures, it would obviously help to be able to specify what we want precisely. Recall that none of these problems would come up if a system were optimizing what we wanted directly, rather than optimizing a proxy.

Unfortunately, this is hard; so can the AI system we're building help us with this? More generally, can a successor agent help its predecessor solve this, leveraging its intellectual advantages to figure out the predecessor's goals?

AIXI learns what to do through a reward signal which it gets from the environment. We can imagine that AIXI's programmers have a button which they press when AIXI does something they like, allowing them to use AIXI's intelligence to solve the value loading problem for them.

The problem with this is that AIXI will apply its intelligence to the problem of taking control of the reward button, since ``control the reward signal'' is an even better reward-maximizing strategy than ``figure out what the reward administrator wants''. This is the problem of \textbf{wireheading} \citep{Bostrom:2014:superintelligence,Amodei:2016:problems}.

This kind of behavior is potentially very difficult to anticipate. The system may deceptively behave as intended during training, planning to take control after deployment. This is the scenario \citet{Bostrom:2014:superintelligence} calls a ``treacherous turn''.

We could perhaps build the reward button \emph{into} the agent, as a black box which issues rewards based on what is occurring. The box could be an \textcolor{g}{intelligent sub-agent} in its own right, which figures out what rewards humans would want to give. The box could even defend itself by issuing punishments for actions aimed at modifying the box.

In the end, however, if the agent understands the situation, it will be motivated to take control anyway. What we want is a solution that still works if we're delegating to a smarter, more capable system.

If the agent is told to get high output from ``the button'' or ``the box'', then it will be motivated to hack those things. However, if you run the expected outcomes of plans through the actual reward-issuing box, then plans to hack the box are evaluated by the box itself, which won't find the idea appealing.

Daniel \citet{Dewey:2011:learning} calls the second sort of agent an \emph{observation-utility maximizer}.\footnote{Others have included observation-utility agents within a more general notion of reinforcement learning.} We find it very interesting to observe that one can try a wide variety of strategies to prevent an advanced RL agent from wireheading, but the agent keeps working against each one; then, one makes the shift to observation-utility agents and the problem vanishes.

However, we still have the problem of specifying the goal $U$. Daniel Dewey points out that observation-utility agents can still use learning to approximate $U$ over time; we just can't treat $U$ as a black box. An RL agent tries to learn to predict the reward function, whereas an observation-utility agent uses estimated utility functions from a human-specified value-learning prior.

However, it remains difficult to specify a learning process which doesn't lead to other problems. For example, if we are trying to learn what humans want, how do we robustly \textcolor{o}{identify ``humans'' in the world}? Merely statistically decent object recognition could lead back to wireheading, as described by \citet{Soares:2015:value}.

Even if we successfully solve that problem, the agent might correctly locate value in the human, but might still be motivated to change human values to be easier to satisfy. Suppose, for example, that there is a drug which modifies human preferences to only care about using the drug. An observation-utility agent could be motivated to give humans that drug in order to make its job easier. This is called the \emph{human manipulation} problem \citep{Soares:2015:corrigibility,Bostrom:2014:superintelligence}.

Anything marked as the true repository of value gets hacked. We might think of this as a case of extremal Goodhart, where the typical action falls in a ``non-wireheading, non-manipulation'' cluster, but options scoring extremely well on a given proxy tend to fall in a cluster of ``wireheading or manipulation'' outliers. Or we might think of this as a case of causal Goodhart, where intervening breaks an empirical correlation between the value loading process we're using and the desired behavior. Whether this is one of the four types of Goodharting, or a fifth, or something all its own, it seems like a theme.\footnote{It might be useful to analyze this as a kind of use/mention violation---the utility function needs to be used, but something goes wrong when it is referenced indirectly. However, this analysis doesn't obviously point toward a way to address the problem.}

The challenge, then, is to create \emph{stable pointers to what we value}: an indirect reference to values not directly available to be optimized, which doesn't thereby encourage hacking the repository of value \citep{Demski:2017:stable}.

\label{rd5} One important point is made by \citet{Everitt:2017:corrupted} in ``Reinforcement Learning with a Corrupted Reward Channel'': the way that one sets up the feedback loop makes an enormous difference. Everitt et al. draw the following picture: 

\begin{center}
  \includegraphics[width=0.9\textwidth]{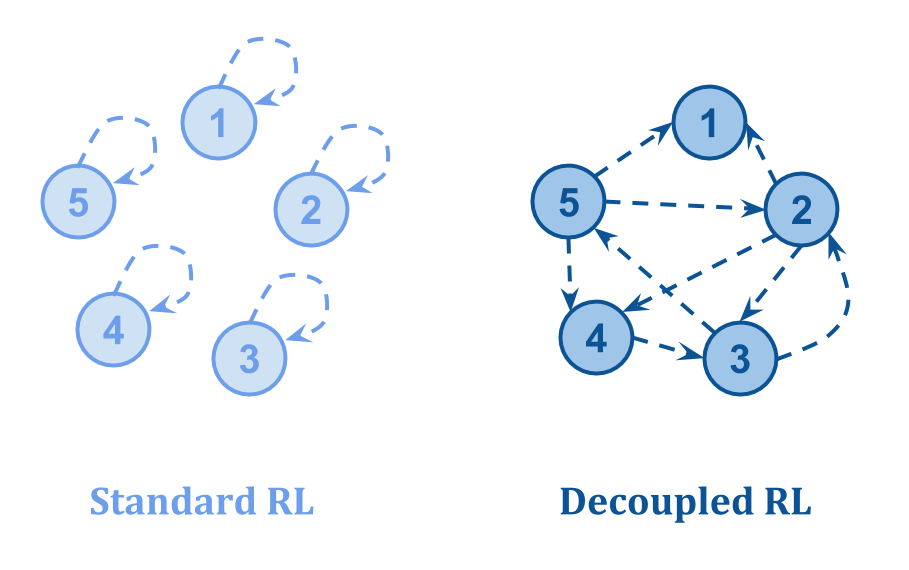}
\end{center}

\noindent In \emph{standard RL}, the feedback about the value of a state comes from the state itself, so corrupt states can be ``self-aggrandizing''. In \emph{decoupled RL}, the feedback about the quality of a state comes from some other state, making it possible to learn correct values even when some feedback is corrupt.

In some sense, the challenge is to put the original, small agent in the feedback loop in the right way. However, the problems with updateless reasoning mentioned earlier make this hard; the original agent doesn't know enough. 

One way to try to address this is through \emph{intelligence amplification}: try to turn the original agent into a more capable one with the same values, rather than creating a successor agent from scratch and trying to get value loading right. 

For example, \citet{Christiano:2018:amplifying} propose an approach in which the small agent is simulated many times in a large tree, which can perform complex computations by \textcolor{g}{splitting problems into parts}.

However, this is still fairly demanding for the small agent. It doesn't just need to know how to break problems down into more tractable pieces; it also needs to know how to do so without giving rise to malign subcomputations. For example, since the small agent can use the copies of itself to acquire a large amount of computational power, it could easily attempt a brute-force search for solutions that ends up running afoul of Goodhart's law.

This issue is the subject of the next section: \textcolor{g}{\emph{subsystem alignment}}.

\section{Subsystem alignment}\label{sec:sa}

You want to figure something out, but you don't know how to do that yet. You have to somehow break up the task into sub-computations. There is no atomic act of ``thinking''; intelligence must be built up of non-intelligent parts. 

The agent being made of parts is part of what made \textcolor{r}{counterfactuals} hard, since the agent may have to reason about impossible configurations of those parts. Being made of parts is what makes \textcolor{b}{self-reasoning and self-modification} even possible.

What we're primarily going to discuss in this section, though, is another problem: when the agent is made of parts, there could be \textcolor{r}{adversaries} not just in the external environment, but inside the agent as well.\footnote{See \citet{Yudkowsky:2017:nonadversarial} for a discussion of ``non-adversarial'' AI as a design goal, and some of the associated challenges.}

This cluster of problems is \textcolor{g}{\textbf{Subsystem Alignment}}: ensuring that subsystems are not working at cross purposes; avoiding subprocesses optimizing for unintended goals. 
\begin{itemize}
 \tightlist
  \item benign induction
  \item benign optimization
  \item transparency
  \item mesa-optimizers
\end{itemize}

\subsection{Robustness to relative scale}
Consider this straw agent design: 

\vspace{6mm}

\begin{center}
  \includegraphics[width=0.9\textwidth, cfbox=gray]{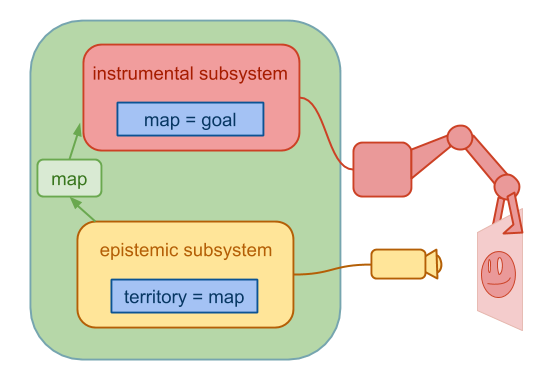}
\end{center}

\vspace{6mm}

\noindent The epistemic subsystem just wants accurate beliefs. The instrumental subsystem uses those beliefs to track how well it is doing. If the instrumental subsystem gets too capable relative to the epistemic subsystem, it may decide to try to fool the epistemic subsystem, as depicted.\footnote{If the epistemic subsystem gets too strong, that could also possibly yield bad outcomes.}

This agent design treats the system's epistemic and instrumental subsystems as discrete agents with goals of their own, which is not particularly realistic. However, we saw in the section on \textcolor{b}{wireheading} that the problem of subsystems working at cross purposes is hard to avoid. And this is a harder problem if we didn't intentionally build the relevant subsystems.

One reason to avoid booting up sub-agents who want different things is that we want \textbf{robustness to relative scale} \citep{Garrabrant:2018:robustness}. 

An approach is \emph{robust to scale} if it still works, or fails gracefully, as you scale capabilities. There are three types: \textcolor{b}{robustness to scaling up}; \textcolor{o}{robustness to scaling down}; and \textcolor{g}{robustness to relative scale}.

\begin{itemize}
  \item \emph{Robustness to scaling up} means that your system doesn't stop behaving well if it gets better at optimizing. One way to check this is to think about what would happen if the function the AI optimizes were actually maximized \citep{Yudkowsky:2015:omnipotence}. Think \textcolor{b}{Goodhart's law}.
  \item \emph{Robustness to scaling down} means that your system still works if made \textcolor{o}{less powerful}. Of course, it may stop being useful; but it should fail safely and without unnecessary costs.

  Your system might work if it can exactly maximize some function, but is it safe if you approximate? For example, maybe a system is safe if it can learn human values very precisely, but approximation makes it increasingly misaligned.

  \item \emph{Robustness to relative scale} means that your design does not rely on the agent's subsystems being similarly powerful. For example, GAN (Generative Adversarial Network) training can fail if one sub-network gets too strong, because there's no longer any training signal \citep{Goodfellow:2014:gan}.
\end{itemize}
\noindent Lack of robustness to scale isn't necessarily something which kills a proposal, but it is something to be aware of; lacking robustness to scale, you need strong reason to think you're at the right scale. 

Robustness to relative scale is particularly important for subsystem alignment. An agent with intelligent sub-parts should not rely on being able to outsmart them, unless we have a strong account of why this is always possible. 

\subsection{Subgoals, pointers, and search}

The big-picture moral: aim to have a unified system that doesn't work at cross purposes to itself.

Why would anyone make an agent with parts fighting against one another? There are three relatively obvious reasons: \emph{subgoals}, \emph{pointers}, and \emph{search}.

Splitting up a task into \textbf{subgoals} may be the only way to efficiently find a solution. However, a subgoal computation should not completely forget the big picture. An agent designed to build houses should not boot up a sub-agent that cares only about building stairs.

One intuitive desideratum is that although subsystems need to have their own goals in order to decompose problems into parts, the subgoals need to ``\textcolor{b}{point back}'' robustly to the main goal. A house-building agent might spin up a subsystem that cares only about stairs, but only cares about stairs in the context of \emph{houses}.

However, we need to achieve this in a way that does not just amount to our house-building system having a second house-building system inside its head. This brings us to the next item.

\vspace{4mm}

\textbf{Pointers:} It may be difficult for subsystems to carry the \textcolor{o}{whole-system} goal around with them, since they need to be \emph{reducing} the problem. However, this kind of indirection seems to encourage situations in which different subsystems' incentives are misaligned. 

As we saw in the case of the agent with epistemic and instrumental subsystems, as soon as we start optimizing some sort of \emph{expectation}, rather than directly getting feedback about what we're doing on the metric that is actually important, we may create perverse incentives---i.e., we are vulnerable to Goodhart's law.

How do we ask a subsystem to ``do X'' as opposed to ``convince the wider system that I'm doing X'', without passing along the entire overarching goal-system? 

This is similar to the way we wanted \textcolor{b}{successor agents} to robustly point at values, since it is too hard to write values down. However, in this case, learning the values of the larger agent would not make any sense either; subsystems and subgoals need to be \textcolor{o}{\emph{smaller}}.

\vspace{4mm}

It might not be that difficult to solve subsystem alignment for subsystems which humans entirely design, or subgoals which an AI explicitly spins up. If you know how to avoid misalignment by design \textcolor{b}{and robustly delegate your goals}, both problems seem solvable.

However, it does not seem possible to design all subsystems so explicitly. At some point in solving a problem, you have split it up as much as you know how to and must rely on some trial and error.

This brings us to the third reason subsystems might be optimizing different things, \textbf{search}: solving a problem by looking through a rich space of possibilities, a space which may itself contain misaligned subsystems. 

Machine learning researchers are quite familiar with the fact that it is easier to write a program which finds a high-performance machine translation system for you than to directly write one yourself. In the long run, this process can go one step further. For a rich enough problem and an impressive enough search process, the solutions found via search might themselves be intelligently optimizing something.

This might happen by accident, or be purposefully engineered as a strategy for solving difficult problems. Either way, it stands a good chance of exacerbating Goodhart-type problems---you now effectively have two chances for misalignment, where you previously had one. This problem is noted by \citet{Yudkowsky:2016:daemon} and described in detail in Hubinger et al.'s \citeyear{Hubinger:2019:mesa} ``Risks from Learned Optimization in Advanced Machine Learning Systems''.

Let us call the original search process the ``base optimizer'', and the search process found via search a ``mesa-optimizer''.

``Mesa'' is the opposite of ``meta''. Whereas a ``meta-optimizer'' is an optimizer designed to produce a new optimizer, a ``mesa-optimizer'' is any optimizer generated by the original optimizer—whether or not the programmers \emph{wanted} their base optimizer to be searching for new optimizers.

``Optimization'' and ``search'' are ambiguous terms. We will think of them as any algorithm which can be naturally interpreted as doing significant computational work to ``find'' an object that scores highly on some objective function.

The \textcolor{b}{objective function} of the base optimizer is not necessarily the same as that of the mesa-optimizer. If the base optimizer wants to make pizza, the new optimizer may enjoy kneading dough, chopping ingredients, et cetera.

The new optimizer’s objective function must be \emph{helpful} for the base objective, at least in the examples the base optimizer is checking. Otherwise, the mesa-optimizer would not have been selected. However, the mesa-optimizer must reduce the problem somehow; there is no point to its running the exact same search. So it seems like its objectives will tend to be like good heuristics; easier to optimize, but different from the base objective in general.

Why might a difference between base objectives and mesa-objectives be concerning, if the new optimizer is scoring highly on the base objective anyway? The concern comes from the interplay between the inner objective and what's really wanted. Even if we get \textcolor{b}{value specification} exactly right, there will always be some \emph{distributional shift} between the training set and deployment. \footnote{See \citet{Amodei:2016:problems} for further discussion of distributional shift.}

Distributional shifts which would be small in ordinary cases may make a big difference to a capable mesa-optimizer, which may observe the slight difference and figure out how to capitalize on it for its own objective.

Indeed, to even use the term ``distributional shift'' seems wrong in the context of embedded agency. The world is not i.i.d. The analog of ``no distributional shift'' would be to have an \textcolor{o}{exact model of the whole future relevant to what you want to optimize}, and the ability to \textcolor{o}{run it over and over} during training. So we need to deal with massive ``distributional shift''.

We may also want to optimize for things that are not exactly what we want. The obvious way to avoid agents that pursue subgoals at the cost of the overall goal is to have the subsystems not be agentic. Just search over a bunch of ways to make stairs---don't make something that cares about stairs. The problem is then that powerful mesa-optimizers are optimizing something the base system doesn’t care about, and that the mesa-optimizers will have a convergent incentive to be agentic \citep{Branwen:2016,Omohundro:2008}.

\subsection{Treacherous turns}

Search processes that are powerful enough to locate other smart search processes raise another problem: the mesa-optimizer may become aware of the base optimizer, in which case it might start explicitly trying to do well on the base objective function purely in order to be kept around, while looking for any signs that it has left training and can stop pretending. This creates a version of Bostrom's \citeyearpar{Bostrom:2014:superintelligence} ``treacherous turn''.

This is the same story we saw in \textcolor{r}{adversarial} \textcolor{b}{Goodhart}: there is something agentic in the search space, which responds to our choice of proxy in a way which makes our proxy a bad one.

One might try to avoid treacherous turns by repeatedly simulating ``end of training, time for deployment'' during training, in a sort of nested-dream setup. The goal is to test enough layers of simulation that you can conclude there will not be a treacherous turn in the real deployment case.

Unfortunately, convergence for this kind of learning is going to be poor. Ordinarily in machine learning, good performance means good average-case performance. But a treacherous turn is an ``error'' which can be carefully placed to do the most damage. We want to ensure that this doesn't happen.

The problem is, in part, that some outputs are much more important than others. Deployment is more important than training, and certain critical or vulnerable moments during deployment will be especially important. We want to be particularly sure to get important things right, rather than just getting low average loss.

But we can't solve this by telling the system what's important. Indeed, it seems we hope it can't figure that out---we are banking on being able to generalize from performance on less-important cases to more-important cases. This is why research into ML techniques which avoid rare catastrophe (or ``traps'') is relevant to the problem of making sure mesa-optimizers are aligned with base optimizers \citep{Christiano:2016:catastrophes,Christiano:2018:worstcase}.

It is difficult to trust arbitrary code---which is what models from rich model classes are---based only on empirical testing. Consider a highly simplified problem: we want to find a program which only ever outputs $1$. $0$ is a catastrophic failure. If we could examine the code ourselves, this problem would be easy. But the output of machine learning is often difficult to analyze; so let's suppose that we can't understand code at all.

Now, in some sense, we can trust simpler functions more. A short piece of code is less likely to contain a hard-coded exception. Let's quantify that.

Consider the set of all programs of length $L$. Some programs $p$ will print $1$ for a long time, but then print $0$. We are trying to avoid that.

Call the time-to-first-zero $W_{p}$. $W_{p} = \infty$ if the program $p$ is trustworthy, i.e., if it never outputs $0$. The highest finite $W_{p}$ out of all length-$L$ programs is a form of the Busy Beaver function, so I will refer to it as $BB(L)$. If we wanted to be completely sure that a random program of length $L$ were trustworthy, we would need to observe $BB(L)$ ones from that program.

Now, a fact about the Busy Beaver function is that $BB(n)$ grows faster than any computable function. So this kind of empirical trust-building takes uncomputably long to find the truth, in the worst case.

What about the average case?

If we suppose all the other length-$L$ programs are easy cases, there are exponentially many length-$L$ programs, so the average is $BB(L) \ / \ \mathrm{exp}(L)$. But exponentials are computable. So $BB(L) \ / \ \mathrm{exp}(L)$ still grows faster than any computable function.
Hence while using short programs gives us some confidence in theory, the difficulty of forming generalized conclusions about behavior grows extremely quickly as a function of length.

If length restrictions aren't so practical, perhaps restricting computational complexity can help us? Intuitively, a mesa-optimizer needs time to think in order to successfully execute a treacherous turn. As such, a program which arrives at conclusions more quickly might be more trustworthy.

However, restricting complexity class unfortunately doesn't get around Busy-Beaver-type behavior. Strategies that wait a long time before outputting $0$ can be slowed down even further with only slightly longer program length $L$.

\vspace{4mm}

If all of these problems seem too hypothetical, consider the evolution of life on Earth. Evolution can be thought of as a reproductive fitness maximizer.\footnote{Evolution can actually be thought of as an optimizer for many things, or as no optimizer at all, but this doesn't matter. The point is that if an agent wanted to maximize reproductive fitness, it might use a system that looked like evolution.} Intelligent organisms are mesa-optimizers of evolution. Although the drives of intelligent organisms are certainly correlated with reproductive fitness, organisms want all sorts of things. There are even mesa-optimizers who have come to understand evolution, and indeed to manipulate it at times.

Powerful and misaligned mesa-optimizers appear to be a real possibility, then, at least with enough processing power. Problems seem to arise because one is trying to solve a problem which one doesn't yet know how to solve by searching over a \textcolor{o}{large} space and hoping ``someone'' can solve it. 

If the source of the issue is the solution of problems by massive search, perhaps we should look for different ways to solve problems. Perhaps we should solve problems by figuring things out. But how do we solve problems which we don't yet know how to solve other than by trying things?

\vspace{6mm}

Let's take a step back. 

\textcolor{o}{Embedded world-models} is about how to think at all, as an embedded agent; \textcolor{r}{decision theory} is about how to act. \textcolor{b}{Robust delegation} is about building trustworthy successors and helpers. \textcolor{g}{Subsystem alignment} is about building \emph{one} agent out of trustworthy \emph{parts}.

The problem is that:

\begin{itemize}
 \tightlist
  \item We don't know how to think about environments when we're \textcolor{o}{smaller}.
  \item To the extent we \emph{can} do that, we don't know how to think about \textcolor{r}{consequences of actions} in those environments.
  \item Even when we can do that, we don't know how to think about \textcolor{b}{what we \emph{want}}.
  \item Even when we have none of these problems, we don't know how to reliably \textcolor{g}{output actions} which get us what we want!
\end{itemize}

\vspace{4mm}

\section{Concluding thoughts}\label{sec:ec}

We described an embedded agent, Emmy, and said that we don't understand how she evaluates her options, models the world, models herself, or decomposes and solves problems. 

One of the main reasons researchers have recently advocated for work on these kinds of problems is \emph{artificial intelligence risk}, as in \citet{Soares:2017:ensuring} and \citet{Bostrom:2014:superintelligence}. AI researchers want to build machines that can solve problems in the general-purpose fashion of a human, and dualism isn't a realistic framework for thinking about such systems. In particular, it's an approximation that's especially prone to breaking down as AI systems become smarter.

We care about basic conceptual puzzles which we think researchers need to figure out in order to achieve confidence in future AI systems, and in order to not have to rely heavily on brute-force search or trial and error. Our hope is that by improving our understanding of embedded agency, we can help make it the case that future developers of general AI systems are a better position to understand their systems, analyze their internal properties, and be confident in their future behavior.

But the arguments for why we may or may not need particular conceptual insights in AI are fairly long, and we haven't attempted to wade into the details of this debate here. We have been considering a particular set of research directions as an \emph{intellectual puzzle}, and not as an instrumental strategy.

These research directions above are largely a refactoring of the problems described in Soares and Fallenstein's (\citeyear{Soares:2015:align}) ``Agent Foundations'' technical agenda. But whereas Soares and Fallenstein framed these issues as ``Here are various things it would be valuable to understand about aligning AI systems'', our framing is more curiosity-oriented: ``Here is a central mystery about how the universe works, and here are a bunch of sub-mysteries providing avenues of attack on the central mystery''. It is not a coincidence that the problem sets overlap, since both sets were generated in the first place by considering which aspects of real-world optimization seemed most conceptually confusing.

One downside of discussing these problems as instrumental strategies is that it can lead to some misunderstandings about why we think this kind of work is important. While employing the ``instrumental strategies'' lens, it is tempting to draw a direct line from a given research problem to a given safety concern. But we are not imagining real-world embedded systems being ``bad at counterfactuals'' or ``bad at world-modeling'' and this somehow causing problems if we don't figure out what is wrong with current models of rational agency. It's certainly not that we're imagining future AI systems being written in second-order logic. As in Yudkowsky's (\citeyear{Yudkowsky:2018:rocket}) ``rocket alignment'' analogy, we're not trying at all to draw direct lines between research problems and specific AI failure modes in most cases.

Our thought on this issue is rather that we seem to be working with the wrong basic concepts today when we try to think about what agency is, as seen by the fact that these concepts don't transfer well to the more realistic embedded framework. If AI developers in the future are still working with these confused and incomplete basic concepts as they try to actually build powerful real-world optimizers, that seems like a bad position to be in. And it appears unlikely to us that the research community will resolve all of these conceptual difficulties by default in the course of just trying to develop more capable systems.\footnote{Evolution certainly managed to build human brains without ``understanding'' any of this, via brute-force search.}

Embedded agency is our way of trying to point at what we think is a very important and central puzzle concerning agency and intelligence---one that we find confusing, and one where we think future researchers risk running into confusions as well.

There is also a significant amount of excellent AI alignment research that is being done with an eye toward more direct applications. In this context, we think of work on ``embedded agency'' or ``Agent Foundations'' as having a different type signature from most other AI alignment research, analogous to the difference between science and engineering. We think of ``Agent Foundations'' research as more like science: more reliant on forward-chaining from curiosity and confusion, rather than backward-chaining from concrete system requirements. Roughly speaking, our goal in working on embedded agency is to build up relevant insights and background understanding, until we collect enough that the alignment problem is more manageable.

Intellectual curiosity isn't the ultimate reason we privilege these research directions. But on our view, there are some \emph{practical} advantages to orienting toward research questions from a place of curiosity at times, as opposed to \emph{only applying the ``practical impact'' lens} to how we think about AI.

When we apply the curiosity lens to the world, we orient toward the sources of confusion preventing us from seeing clearly; the blank spots in our map, the flaws in our lens. It encourages re-checking assumptions and attending to blind spots, which is helpful as a psychological counterpoint to our ``instrumental strategy'' lens---the latter being more vulnerable to the urge to lean on whatever shaky premises we have on hand so we can get to more solidity and closure in our early thinking.

\emph{Embedded agency} is an organizing theme behind most, if not all, of our big curiosities. It seems like a central mystery underlying many concrete difficulties.

\clearpage

\printbibliography

\end{document}